\documentclass{article}

\usepackage{arxiv}

\usepackage[utf8]{inputenc} 
\usepackage[T1]{fontenc}    
\usepackage{hyperref}       
\usepackage{url}            
\usepackage{booktabs}       
\usepackage{amsfonts}       
\usepackage{nicefrac}       
\usepackage{microtype}      
\usepackage[ruled,vlined]{algorithm2e}    
\usepackage{amsmath}
\newcommand\norm[1]{\left\lVert#1\right\rVert}  
\usepackage{graphicx}       
\usepackage{subfig}         
\usepackage{comment}        

\title{An algorithmic Introduction to Clustering}

\author{
  Bernardo Gonzalez \\
  Department of Computer Science and Engineering\\
  UCSC\\
  \texttt{beaugonz@ucsc.edu} \\
}

\begin{document}
\maketitle

The purpose of this document is to provide an easy introductory guide to clustering algorithms. Basic knowledge and exposure to probability (random variables, conditional probability, Bayes' theorem, independence, Gaussian distribution), matrix calculus (matrix and vector derivatives), linear algebra and analysis of algorithm (specifically, time complexity analysis) is assumed.

Starting with \textit{Gaussian Mixture Models} (GMM), this guide will visit different algorithms like the well-known \textit{$k$-means}, \textit{DBSCAN} and \textit{Spectral Clustering} (SC) algorithms, in a connected and hopefully understandable way for the reader. The first three sections (Introduction, GMM and $k$-means) are based on \cite{bishop}. The fourth section (SC) is based on \cite{dhillon} and \cite{scholkopf}. Fifth section (DBSCAN) is based on \cite{schubert}. Sixth section (\textit{Mean Shift}) is, to the best of author's knowledge, original work. Seventh section is dedicated to conclusions and future work.

Traditionally, these five algorithms are considered completely unrelated and they are considered members of different families of clustering algorithms:

\begin{itemize}
    \item \textit{Model-based algorithms}: the data are viewed as coming from a mixture of probability distributions, each of which represents a different cluster. A Gaussian Mixture Model is considered a member of this family
    \item \textit{Centroid-based algorithms}: any data point in a cluster is represented by the central vector of that cluster, which need not be a part of the dataset taken. $k$-means is considered a member of this family
    \item \textit{Graph-based algorithms}: the data is represented using a graph, and the clustering procedure leverage \textit{Graph theory} tools to create a partition of this graph. Spectral Clustering is considered a member of this family
    \item \textit{Density-based algorithms}: this approach is capable of finding arbitrarily shaped clusters, where clusters are defined as dense regions separated by low-density regions. DBSCAN and Mean Shift are considered members of this family
\end{itemize}

In this document, we will try to present a more unified view of clustering, by identifying the relationships between the algorithms mentioned. Some of the results are not new, but they are presented in a cleaner, simpler and more concise way. To the best of author's knowledge, the interpretation of DBSCAN as a climbing procedure, which introduces a theoretical connection between DBSCAN and Mean shift, is a novel result.

\begin{figure}[h]
    \centering
    \includegraphics[width=0.8\textwidth]{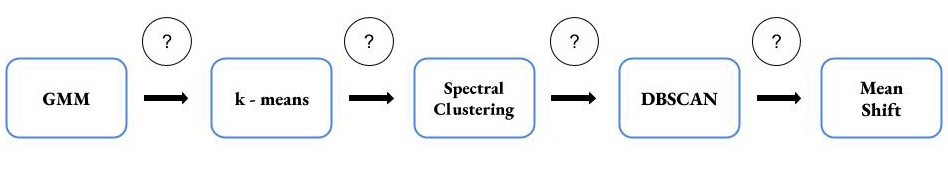}
    \caption{What is the relation between these clustering algorihtms?}
    \label{fig:algs}
\end{figure}

\section{Introduction}

The general version of the \textit{clustering problem} is the following: given a set $X$ of $n$ i.i.d. (independent and identically distributed) points $x_i \in \mathcal{S}, i = 1, 2, ..., n$, and a distance function $dist: \mathcal{S} \times \mathcal{S} \to \mathbb{R}_+$, cluster them in such a way that points in the same group are less distant to each other than to those in other clusters. In order to keep things simple, for the rest of this document we will assume that $\mathcal{S} = \mathbb{R}^d$, but keep in mind that most of the algorithms presented here can be generalized to spaces different than $\mathbb{R}^d$. Also, a similarity measure $sim: \mathcal{S} \times \mathcal{S} \to \mathbb{R}$ could be used instead of a distance function. In this case, however, the goal is to cluster the points in such a way that points in the same group are more similar to each other than to those in other clusters. 

\subsection{Maximum Likelihood for the Gaussian distribution}

Let $x \in \mathbb{R}^d$ be a $d$-dimensional random vector. A Gaussian distribution (also known as Normal distribution) with mean vector $\mu \in \mathbb{R}^d$ and covariance matrix $\Sigma \in \mathbb{R}^{d \times d}$ can be written in the form:

\begin{equation}
\mathcal{N}(x|\mu, \Sigma) = \frac{1}{(2\pi)^{\frac{d}{2}}}\frac{1}{|\Sigma|^{\frac{1}{2}}}\exp{\Big(-\frac{1}{2}(x - \mu)^T\Sigma^{-1}(x - \mu)\Big)}
\end{equation}

where $|\Sigma|$ denotes the determinant of $\Sigma$ and $x^T$ is the transpose of vector $x$.

Given a dataset $X$ composed by $n$ i.i.d. data points $x_1, x_2, ..., x_n$, where $\forall i, x_i \in \mathbb{R}^d$, we can use a Gaussian Distribution as a model to fit this dataset. A common criterion for fitting the parameters $\mu$ and $\Sigma$ of the Gaussian distribution given an observed dataset is to find the values that maximize the likelihood function:

\begin{equation}
p(X|\mu, \Sigma) = \prod^n_{i=1} \mathcal{N}(x_i|\mu, \Sigma) = \prod^n_{i=1} \frac{1}{(2\pi)^{\frac{d}{2}}}\frac{1}{|\Sigma|^{\frac{1}{2}}}\exp\Big(-\frac{1}{2}(x_i - \mu)^T\Sigma^{-1}(x_i - \mu)\Big)
\end{equation}

This procedure is known as \textit{Maximum Likelihood Estimation (MLE)}. Given the form of the Gaussian distribution, it is more convenient to maximize the $\log$ of the likelihood function (maximizing this is equivalent because the logarithmic function is a monotonically increasing function). Taking $\log$ on both sides

\begin{equation}
\log p(X|\mu, \Sigma) = -\frac{nd}{2}\log(2\pi) - \frac{n}{2}\log|\Sigma| - \frac{1}{2}\sum^n_{i=1}(x_i - \mu)^T\Sigma^{-1}(x_i - \mu)
\end{equation}

Maximizing with respect to $\mu$ (i.e., taking the derivative of $\log p(X|\mu, \Sigma)$ w.r.t. $\mu$ and equalizing to zero), we obtain

\begin{equation}
\mu_{MLE} = \frac{1}{n}\sum^n_{i=1}x_i
\end{equation}

Similarly, maximizing w.r.t. $\Sigma$

\begin{equation}
\Sigma_{MLE} = \frac{1}{n}\sum^n_{i=1}(x_i - \mu_{MLE})(x_i - \mu_{MLE})^T
\end{equation}

As we can see, fitting one Gaussian distribution to a dataset $X$ using maximum likelihood is a simple procedure, as shown in algorithm \ref{algo:MLGD}. The complexity of this algorithm is $O(nd^2)$, due the computation of matrix $\Sigma_{MLE}$.

\begin{algorithm}
\DontPrintSemicolon 
\KwIn{Dataset $X = \{ x_1, x_2, ..., x_n\}$}
\KwOut{Learned values for $\mu$ and $\Sigma$}
Estimate the parameters $\mu$ and $\Sigma$:

\begin{align*}
\mu_{MLE} &\gets \frac{1}{n}\sum^n_{i=1}x_i \\
\Sigma_{MLE} &\gets \frac{1}{n}\sum^n_{i=1}(x_i - \mu_{MLE})(x_i - \mu_{MLE})^T \\
\end{align*}
\Return{$\mu_{MLE}, \Sigma_{MLE}$}\;
\caption{Maximum Likelihood for one Gaussian Distribution}
\label{algo:MLGD}
\end{algorithm}

\begin{figure}%
    \centering
    \subfloat[Dataset with 500 points in two dimensions]{{\includegraphics[width=0.45\textwidth]{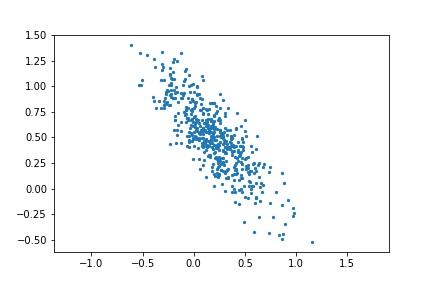} }}%
    \qquad
    \subfloat[Gaussian model learned from data using MLE]{{\includegraphics[width=0.45\textwidth]{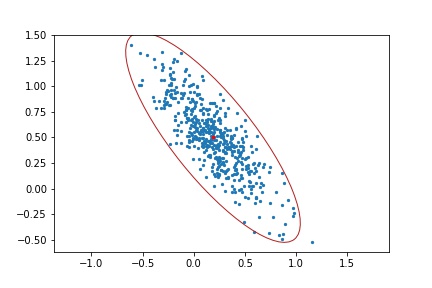} }}%
    \caption{Fitting one Gaussian distribution to 500 data points. See text for details}%
    \label{fig:dataOneGauss}%
\end{figure}

Figure \ref{fig:dataOneGauss}(a) shows a dataset with 500 points in two dimensions. Figure \ref{fig:dataOneGauss}(b) shows the same dataset and the learned configuration for the Gaussian model. The mean and the three standard-deviation contour for this model are shown as a red point and a red ellipse, respectively.

\newpage
\subsection{Sampling from the learned model}

\begin{figure}[h]
    \centering
    \includegraphics[width=0.5\textwidth]{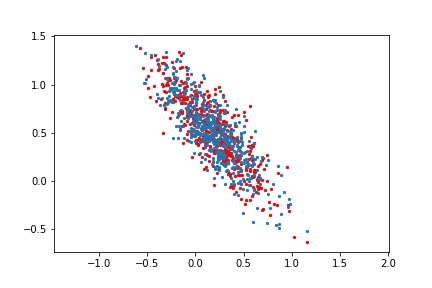}
    \caption{Points sampled from learned distribution (in red) along with original dataset (in blue)}
    \label{fig:dataGenOneGauss}
\end{figure}

After fitting the model to the data distribution, we can sample from it and compare how a set of samples looks with respect to the original dataset. Figure \ref{fig:dataGenOneGauss} shows a set of 500 points sampled from the data (in red) along with the 500 original data points (in blue). We can observe a couple of things: the space covered by the sampled points is similar to the space covered by the original data points, and any of the sampled points are "close" to at least one of the original data points. Despite not being a quantitative measure, figure \ref{fig:dataGenOneGauss} allow us to form an intuition of how well the model fitted the data for this two dimensional dataset.

\section{Gaussian Mixture Models (GMM)}

\subsection{Only one Gaussian: not flexible enough}

\begin{figure}%
    \centering
    \subfloat[1500 data points in two dimensions]{{\includegraphics[width=0.25\textwidth]{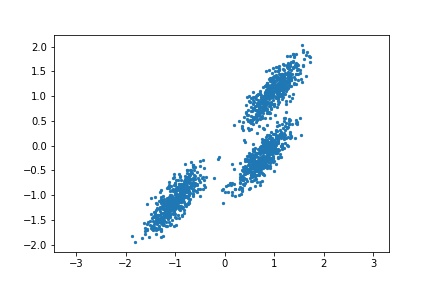} }}%
    \qquad
    \subfloat[Gaussian model learned from data]{{\includegraphics[width=0.25\textwidth]{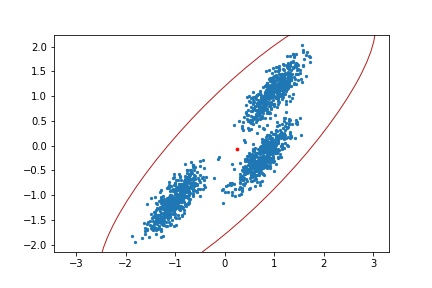} }}%
    \qquad
    \subfloat[Points sampled from learned distribution (in red) along with original dataset (in blue)]{{\includegraphics[width=0.25\textwidth]{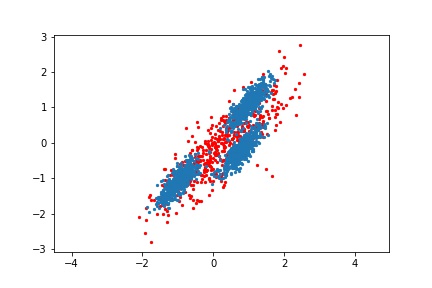} }}%
    \caption{Fitting one Gaussian distribution to 1500 data points}%
    \label{fig:dataOneMultGauss}%
\end{figure}

As we can imagine, there exists datasets for which one Gaussian distribution doesn't offer the flexibility to fit the data points in the dataset to analize. Figure \ref{fig:dataOneMultGauss}(a) shows a dataset with 1500 points in two dimensions. We will denote this dataset as $X_{1500}$. Figure \ref{fig:dataOneMultGauss}(b) shows the dataset $X_{1500}$ and the learned configuration for one Gaussian model. Like in the previous example, the three standard-deviation contour is shown as a red ellipse. However, in this case, we can observe how the model doesn't capture the data distribution as well as in the previous example. Specifically, sampling from the learned distribution generates points which are "far" from any of the original data points, as shown in figure \ref{fig:dataOneMultGauss}(c). Also, we can observe how the sampled points cover parts of the space not covered by the original data points.

\subsection{EM-GMM algorithm}

One way to add more flexibility to our model is to use instead a \textit{Mixture of Gaussians}. A superposition of $k$ Gaussian densities of the form:

\begin{equation} \label{eq:gmm}
p(x) = \sum^k_{c=1}\pi_c\mathcal{N}(x|\mu_c, \Sigma_c)
\end{equation}

is called a mixture of Gaussians. Each Gaussian density $\mathcal{N}(x|\mu_c, \Sigma_c)$ is called a component of the mixture and has its own mean $\mu_c$ and covariance matrix $\Sigma_c$. The parameters $\pi_c \in \mathbb{R}$ are called mixing coefficients, and have two important properties:

\begin{align}
0 \leq \pi_c \leq 1 \label{eq:mix_coeff_prop1} \\
\sum^k_{c=1}\pi_c = 1 \label{eq:mix_coeff_prop2}
\end{align}


We can view the mixing coefficients $\pi_c$ as the prior probability $p(c)$ of picking the $c$-th component. Similarly, we can view the Gaussian densities $\mathcal{N}(x|\mu_c, \Sigma_c)$ as the probability of $x$ conditioned on $c$, $p(x|c)$ (i.e., the likelihood of point $x$ given the $c$-th component). Using this notation, we can express equation \eqref{eq:gmm} as:

\begin{equation}
p(x) = \sum^k_{c=1}p(c)p(x|c)
\end{equation}

From here, we can see that we can compute the posterior probabilities $p(c|x)$ using Bayes' theorem:

\begin{equation} \label{eq:responsibilities}
\lambda_{i,c} \equiv p(c|x_i) =  \frac{p(c)p(x_i|c)}{\sum^k_{j=1}p(j)p(x_i|j)} = \frac{\pi_c\mathcal{N}(x_i|\mu_c, \Sigma_c)}{\sum^k_{j=1}\pi_j\mathcal{N}(x_i|\mu_j, \Sigma_j)}
\end{equation}

Following the notation in \cite{bishop}, we denote the variables $\lambda_{i,c}$ as "responsibilities", and we will use the following notation: $\boldsymbol{\pi} \equiv \{\pi_1, ..., \pi_k\}$, $\boldsymbol{\mu} \equiv \{\mu_1, ..., \mu_k\}$ and $\boldsymbol{\Sigma} \equiv \{\Sigma_1, ..., \Sigma_k\}$. These responsibilities $\lambda_{i,c}$ can be interpreted as a measure of how much the point $x_i$ "belongs" to the $c$-th component, given the current estimates $\boldsymbol{\mu}$ and $\boldsymbol{\Sigma}$. Similar to what we did with one Gaussian distribution, we can try to fit the parameters of this model given an observed dataset $X$ using maximum likelihood. Using the i.i.d. assumption and equation \eqref{eq:gmm} we have

\begin{equation} \label{eq:log_lik_gmm}
\log p(X|\boldsymbol{\pi}, \boldsymbol{\mu}, \boldsymbol{\Sigma}) = \log \Big(\prod^n_{i=1} p(x_i|\boldsymbol{\pi}, \boldsymbol{\mu}, \boldsymbol{\Sigma})\Big) = \sum^n_{i=1}\log\Big(\sum^k_{c=1}\pi_c\mathcal{N}(x_i|\mu_c, \Sigma_c)\Big)
\end{equation}

The situation now is much more complex than with a single Gaussian, due to the presence of the summation inside the logarithm. As a result, the maximum likelihood solution for the parameters of the GMM no longer has a closed-form analytical solution. Maximizing equation \eqref{eq:log_lik_gmm} with respect to $\mu_c$, we obtain

\begin{equation} \label{eq:mu_gmm}
\mu_c = \frac{1}{n_c}\sum^n_{i=1}\lambda_{i,c}x_i
\end{equation}

where $n_c = \sum^n_{i=1}\lambda_{i,c}$. Similarly, maximizing equation \eqref{eq:log_lik_gmm} w.r.t. $\Sigma_c$

\begin{equation} \label{eq:cov_gmm}
\Sigma_c = \frac{1}{n_c}\sum^n_{i=1}\lambda_{i,c}(x_i - \mu_c)(x_i - \mu_c)^T
\end{equation}

Finally, we need to maximize equation \eqref{eq:log_lik_gmm} w.r.t. $\pi_c$. However, in this case we need to take into acount the constraints \eqref{eq:mix_coeff_prop1} and \eqref{eq:mix_coeff_prop2}. This can be achieved using Lagrange multipliers, getting as result

\begin{equation} \label{eq:mc_gmm}
\pi_c = \frac{n_c}{n}
\end{equation}

Despite looking as closed-form solutions, equations \eqref{eq:mu_gmm},\eqref{eq:cov_gmm} and \eqref{eq:mc_gmm} are not such, because all of them are functions of the responsibilities $\lambda_{i,c}$. However, these responsibilities depends on $\boldsymbol{\pi}, \boldsymbol{\mu}, \boldsymbol{\Sigma}$ in a complex way, as we can see from equation \eqref{eq:responsibilities}.

However, the last three equations suggest a simple iterative algorithm to find a solution to the GMM maximum likelihood problem:

\begin{itemize}
  \item First, we choose some initial values for the means $\mu_c$, matrices $\Sigma_c$ and mixing coefficients $\pi_c$
  \item Then, we alternate between the following two updates, that we shall call the \textbf{E step} (from Expectation) and \textbf{M step} (from Maximization):

    \begin{itemize}
        \item In the \textbf{E step}, we use the current values of $\mu_c$, $\Sigma_c$ and $\pi_c$ to compute the responsibilities $\lambda_{i,c}$ using equation \eqref{eq:responsibilities}
        \item Then, we use these $\lambda_{i,c}$ in the \textbf{M step} to re-estimate the values of $\mu_c$, $\Sigma_c$ and $\pi_c$ using equations \eqref{eq:mu_gmm},\eqref{eq:cov_gmm} and \eqref{eq:mc_gmm}, respectively
    \end{itemize}
    
\end{itemize}

 It is important to highlight that when computing the update for $\Sigma^{new}_c$, we use the updated values of $\mu^{new}_c$. The algorithm iterates until a convergence criteria is met. In practice, the algorithm is deemed to have converged when the change in the log likelihood function , or alternatively in the parameters, falls below some threshold.

This iterative procedure using the \textbf{E step} and \textbf{M step} is an instance of a more general algorithm called \textit{Expectation-Maximization algorithm}. The EM algorithm is an iterative procedure to find maximum likelihood solutions for models having latent variables. It should be emphasized that commonly, the log likelihood function has multiple local maxima, and that the EM algorithm is not guaranteed to find the global maxima. The EM algorithm for Gaussian Mixtures models (EM-GMM algorithm) is shown in algorithm \ref{algo:GMM}.

\begin{figure}[h]%
    \centering
    \subfloat[GMM with three components learned from data]{{\includegraphics[width=0.45\textwidth]{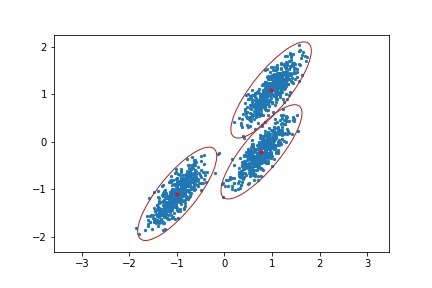} }}%
    \qquad
    \subfloat[Points sampled from GMM (in red) along with original dataset (in blue)]{{\includegraphics[width=0.45\textwidth]{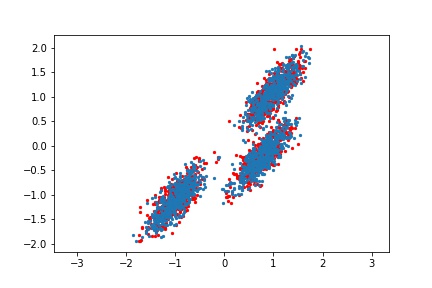} }}%
    \caption{Fitting a GMM with three components to 1500 data points}%
    \label{fig:dataFitMultGauss}%
\end{figure}

EM-GMM is the first clustering algorithm presented in this document: it takes a collection $X$ of $n$ points in a $d$ dimensional space and group them into $k$ different clusters $\Gamma_1, \Gamma_2, ..., \Gamma_k$ using a mixture of Gaussians as a model. Every Gaussian distribution models a cluster $\Gamma_c$. Figure \ref{fig:dataFitMultGauss}(a) shows how a GMM model with three components is able to fit the previous dataset more tightly than the one-Gaussian model. Also, sampling from the GMM model generates points which are more similar to points in the original dataset, as shown in figure \ref{fig:dataFitMultGauss}(b). We can also observe how the space covered by the sampled points is now more similar to the space covered by the original data points.

\begin{algorithm}
\DontPrintSemicolon 
\KwIn{Dataset $X$, number of components $k$, and initial values for every $\pi_c, \mu_c, \Sigma_c$}
\KwOut{Learned values for every $\pi_c, \mu_c, \Sigma_c$ and $\lambda_{i,c}$}
\While{Convergence criterion is not satisfied}{
\textbf{E step}. Evaluate the responsibilities $\lambda_{i,c}$ using the current values of $\pi_c, \mu_c, \Sigma_c$: \\

\begin{equation*}
\lambda_{i,c} \gets \frac{\pi_c\mathcal{N}(x_i|\mu_c, \Sigma_c)}{\sum^k_{j=1}\pi_j\mathcal{N}(x_i|\mu_j, \Sigma_j)}\;
\end{equation*}

\textbf{M step}. Re-estimate the parameters using the current responsibilities $\lambda_{i,c}$:

\begin{align*}
n_c &\gets \sum^n_{i=1}\lambda_{i,c} \\
\mu^{new}_c &\gets \frac{1}{n_c}\sum^n_{i=1}\lambda_{i,c}x_i \\
\Sigma^{new}_c &\gets \frac{1}{n_c}\sum^n_{i=1}\lambda_{i,c}(x_i - \mu^{new}_c)(x_i - \mu^{new}_c)^T \\
\pi^{new}_c &\gets \frac{n_c}{n}\;
\end{align*}
}
\Return{$\boldsymbol{\pi}, \boldsymbol{\mu}, \boldsymbol{\Sigma}, \boldsymbol{\lambda}$}\;
\caption{EM-GMM}
\label{algo:GMM}
\end{algorithm}

\subsection{EM-GMM complexity}

At every iteration, the two updates more computationally intensive are the update of $\lambda_{i,c}$ in \textbf{E step} and the update of the matrices $\Sigma_c$ in the \textbf{M step}.

Specifically, at every iteration, the \textbf{E step} requires the evaluation of $nk$ Gaussian distributions, which involves the computation (for every cluster $\Gamma_c$) of the determinant and the inverse of a matrix of size $d \times d$, both with complexity $O(d^{2.807})$ for practical purposes (using Strassen algorithm for matrix multiplication, however there are algorithms \underline{asymptotically} faster). Also, for every $x_i$, we need to compute a quadratic form w.r.t. every cluster $\Gamma_c$, with complexity $O(d^2)$, therefore resulting in an overall complexity of $O(nd^2k)$. Therefore, the overall complexity for \textbf{E step} at every iteration is $O(nd^2k + d^{2.807}k)$.

Similarly, the update of any of the matrices $\Sigma_c$ requires the computation and addition of $n$ matrices of size $d \times d$, with complexity $O(nd^2)$, therefore resulting in an overall complexity of $O(nd^2k)$ for \textbf{M step} at every iteration.

Therefore, the complexity of the EM-GMM algorithm is $O(nd^2kT + d^{2.807}kT)$, where $T$ denotes the number of iterations. This complexity and other important properties of the EM-GMM algorithm are summarized in table \ref{tab:GMM}.

\begin{table}[h]
\centering
\caption{EM-GMM algorithm}
\begin{tabular}[t]{lc}
\hline
&GMM\\
\hline
Number of hyperparameters & 1 \\
Hyperparameters & number of components $k$  \\
Time complexity & $O(nd^2kT + d^{2.807}kT)$ \\
Outlier detection & No \\
Type of assignment & soft \\
distance/similarity measure & Mahalanobis distance (See $k$-means section for details) \\
\hline
\end{tabular}
\label{tab:GMM}
\end{table}

\subsection{GMM clustering}

A GMM returns what is called a soft assignment for every data point $x_i$: instead of assigning it to a specific cluster $\Gamma_c$, it returns a "responsibility" $\lambda_{i,c}$, expressing how "responsible" is cluster $\Gamma_c$ for data point $x_i$. The set of $\lambda_{i,c}$ for a specific point $x_i$ forms a distribution over the clusters $\Gamma_1, \Gamma_2, ..., \Gamma_k$. On the other hand, a hard assignment is an assignment where every point $x_i$ is assigned to exactly one cluster $\Gamma_c$. 

\begin{figure}[h]
    \centering
    \includegraphics[width=0.5\textwidth]{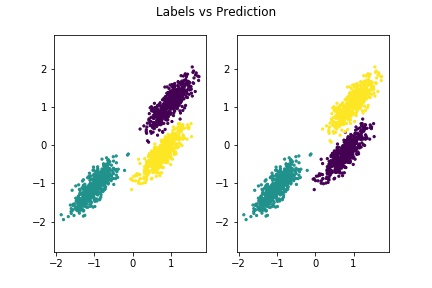}
    \caption{Clustering induced by ground truth (labels) and clustering learnt by GMM}
    \label{fig:dataPredMultGauss}
\end{figure}

One simple way to turn the soft assignment returned by a GMM into a hard assignment is to assign point $x_i$ to the cluster $\Gamma_c$ with larger responsability $\lambda_{i,c}$. Using this rounding method, we can observe in figure \ref{fig:dataPredMultGauss} how the clusters found by the GMM model (right side) compare to the labels from the $X_{1500}$ dataset (left side).

\subsubsection{Clustering quality measures: Adjusted Mutual Information (AMI) and Adjusted Rand Index (ARI)}

In order to present a more quantitative measure of how similar are the labels from the data and the clusters found by the GMM, we introduce two widely-used methods for comparing a pair of clusterings (or more generally, two partitions of a set): the Adjusted Mutual Information (AMI) \cite{hubert} and the Adjusted Rand Index (ARI) \cite{vinh}. AMI is an entropy-based measure that quantifies the overlap between the clusters found and the clusters defined by the labels. Similarly, ARI is a measure that quantifies the number of agreements and disagreements between the clusters found and the clusters defined by the labels. Both scores take the value of 1 when two clusterings are identical. For more details, please refer to \cite{hubert} and \cite{vinh}, respectively.

The AMI and the ARI scores between the labels and the clusters found by the GMM for the dataset shown in figure \ref{fig:dataPredMultGauss} are 0.992 and 0.996, respectively.

\section{$k$-means}

$k$-means is another clustering algorithm, with some similarities with the EM-GMM algorithm (both are iterative algorithms and both need the number of clusters $k$ as an input). Can we derive the $k$-means algorithm from EM-GMM? In the next section we'll observe how we can derive the $k$-means algorithm as a particular limit of the EM-GMM algorithm.

\subsection{From GMM to $k$-means}

Consider a GMM model with all covariance matrices equal to the identity matrix $I$ times a variance parameter $\epsilon$, i.e., $\Sigma_c = \epsilon I$ $\forall c$. Then, for point $x$ and cluster $\Gamma_c$

\begin{equation}
p(x|\mu_c, \Sigma_c) = \frac{1}{(2\pi\epsilon)^{\frac{d}{2}}}\exp{\Big(-\frac{1}{2\epsilon}\norm{x - \mu_c}^2\Big)}
\end{equation}

where we used the identity $\norm{x}^2 = x^Tx$. Lets now consider algorithm \ref{algo:GMM} for a GMM with $k$ Gaussians of this form and treat $\epsilon$ as a fixed constant. In the \textbf{E step}, the responsibilities $\lambda_{i,c}$ for data point $x_i$ are now given by

\begin{equation}
\lambda_{i,c} = \frac{\pi_c\exp{\Big(-\frac{1}{2\epsilon}\norm{x_i - \mu_c}^2\Big)}}{\sum^k_{j=1}\pi_j\exp{\Big(-\frac{1}{2\epsilon}\norm{x_i - \mu_j}^2\Big)}}
\end{equation}

If we consider the limit $\epsilon \to 0$, we observe that in the denominator the term for which $\norm{x_i - \mu_j}^2$ is smallest will go to zero most slowly, and hence the responsibilities $\lambda_{i,c}$ for the data point $x_i$ all go to zero except for term $j$, for which the responsibility $\lambda_{i,j}$ will go to one. Note that this holds independently of the values of $\pi_c$ as long as none of the $\pi_c$ is zero. Therefore, in the limit $\epsilon \to 0$, algorithm \ref{algo:GMM} returns a hard assignment: it assigns every data point $x_i$ to just one specific cluster $\Gamma_c$. It is easy to observe that data points will be assigned to the cluster having the closest mean $\mu_c$. This is the \textbf{E step} in the $k$-means algorithm: assign every point to the cluster with the current closest mean $\mu_c$. Also, instead of using $\lambda_{i,c}$ as the assignments notation, we will follow \cite{bishop} and use $r_{i,c}$ instead as assignments notation in the $k$-means algorithm.

Continuing with the EM-GMM algorithm, lets now observe what changes in the \textbf{M step}. The first thing that we can observe is that the last two equations (updating the covariance matrices $\Sigma^{new}_c$ and $\pi^{new}_c$) are no longer needed: all matrices $\Sigma_c$ are constant and equal, and we no longer need $\pi_c$ for the \textbf{E step}. Also, we can merge the first two equations in \textbf{M step} into one equation which updates the means $\mu_c$ to the mean of the current cluster (after the \textbf{E step} update):

\begin{equation} \label{eq: mean_kmeans}
\mu_c = \frac{\sum^n_{i=1}r_{i,c}x_i}{\sum^n_{i=1}r_{i,c}}
\end{equation}

So, the answer to the question opening this section "Can we derive the $k$-means algorithm from the EM-GMM one?" is yes: using the EM-GMM algorithm with a GMM with all covariance matrices $\Sigma_c = \epsilon I$ and taking the limit when $\epsilon \to 0$, leads to the $k$-means algorithm, shown in algorithm \ref{algo:kmeans}. The consequences of these changes made to the EM-GMM algorithm is that the $k$-means algorithm lose some flexibility w.r.t. a GMM: all points are assigned now to a single cluster, and, because all covariances matrices are diagonal matrices, all data points are assigned to the cluster with the closest mean (i.e., for $k$-means, "closest" means closest w.r.t. to Euclidean distance, while for a GMM, "closest" means closest w.r.t. to Mahalanobis distance \cite{mahalanobis}).

However, the $k$-means algorithm is more efficient than EM-GMM: we do not need to evaluate the probabilities $\mathcal{N}(x_i|\mu_c, \Sigma_c)$ for every data point and every cluster at \textbf{E step}. Also, is not necessary to update the covariance matrices $\Sigma^{new}_c$ at \textbf{M step}. Avoiding these updates reduces significantly the complexity of the algorithm, which is $O(ndkT)$, because at \textbf{E step} we need to compute the distance between every data point $x_i$ and every cluster mean $\mu_c$, at every iteration. This complexity and other important properties of the $k$-means algorithm are summarized in table \ref{tab:kmeans}.

\begin{figure}[h]%
    \centering
    \subfloat[A GMM is flexible enough to fit the data]{{\includegraphics[width=0.45\textwidth]{DataFitMultGDMultGaussDist.jpg} }}%
    \qquad
    \subfloat[$k$-means lacks the flexibility to fit this dataset]{{\includegraphics[width=0.45\textwidth]{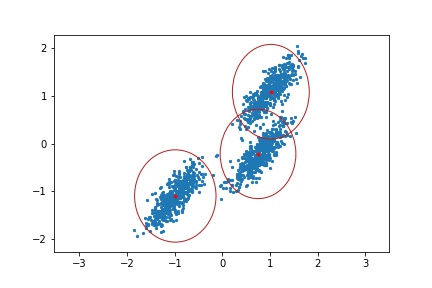} }}%
    \caption{Difference between a GMM and $k$-means}%
    \label{fig:dataFitKMMultGauss}%
\end{figure}

Figure \ref{fig:dataFitKMMultGauss} illustrates the main difference between $k$-means and a GMM. We can observe how $k$-means is not flexible enough to capture the elliptical shape of the data, in contrast to the results obtained using a GMM. The main reason for this is the distance function used by each algorithm, the Mahalanobis distance for GMMs and the Euclidean distance for $k$-means. The circles shown for $k$-means are for illustration purpose only: as mentioned before, $k$-means does not compute the covariance matrices $\Sigma_c$.

\begin{algorithm}
\DontPrintSemicolon 
\KwIn{Dataset $X$, number of clusters $k$, and initial values for every $\mu_c$}
\KwOut{Learned values for every $\mu_c$ and $r_{i,c}$}
\While{Convergence criterion is not satisfied}{
\textbf{E step}. Using the current values of $\mu_c$, assign every point to its closest cluster: \\

\begin{equation*}
r_{i,c} \gets
\begin{cases}
1 & \text{if $c = argmin_j||x_i - \mu_j||^2$}\\
0 & \text{otherwise}
\end{cases}       
\end{equation*}

\textbf{M step}. Re-estimate the cluster means using the current assignations $r_{i,c}$:

\begin{equation*}
\mu_c \gets \frac{\sum^n_{i=1}r_{i,c}x_i}{\sum^n_{i=1}r_{i,c}}
\end{equation*}
}
\Return{$\boldsymbol{\mu}, \boldsymbol{r}$}\;
\caption{$k$-means}
\label{algo:kmeans}
\end{algorithm}

\begin{table}[h]
\centering
\caption{$k$-means algorithm}
\begin{tabular}[t]{lc}
\hline
&$k$-means \\
\hline
Number of hyperparameters & 1 \\
Hyperparameters & number of clusters $k$  \\
Time complexity & $O(ndkT)$ \\
Outlier detection & No \\
Type of assignment & hard \\
distance/similarity measure & Euclidean distance \\
\hline
\end{tabular}
\label{tab:kmeans}
\end{table}

\subsection{$k$-means as an optimization problem}

Before continuing, we'll introduce a different formulation for the $k$-means problem that will be useful in the next section. Given a dataset $X$ with $n$ data points $x_1, x_2, ..., x_n$, the $k$-means algorithm tries to find $k$ points in the space (the means $\mu_c$) and assign every data point $x_i$ to one of these $k$ points such that the distance between $x_i$ and $\mu_c$ is minimized. We can then think of the $k$-means algorithm as an iterative algorithm that returns a local minima for the following optimization problem:

\begin{equation}
\begin{aligned} \label{eq:kmm_opt_prob}
\min_{r_{i,c}, \mu_c} \quad & \sum^k_{c = 1}\sum_{i \in \Gamma_c}\norm{x_i - \mu_c}^2 \\
\textrm{s.t.} \quad & r_{i,c} \in \{0, 1\} \qquad (\forall i = 1, 2, ... n; c = 1, 2, ... k) \\
\end{aligned}
\end{equation}

where we have used the notation $i \in \Gamma_c$ to denote that point $x_i$ belongs to cluster $\Gamma_c$, i.e., $r_{i,c} = 1$ and $r_{i,j} = 0$ $\forall j \neq c$. We'll denote this optimization problem as the vanilla $k$-means problem. A solution for this problem consists of $k$ cluster means $\mu_c$ and a hard assignment $r_{i,c}$ for every data point $x_i$. The $k$-means algorithm returns a local minima for this optimization problem using the EM algorithm.

\section{Spectral Clustering (SC)}

Spectral Clustering is a family of algorithms that use the eigenvectors of a matrix derived from a dataset $X$ to cluster the data points in $X$. Each specific algorithm use the eigenvectors in a slightly different way. In this document, we will focus on the spectral algorithm of Ng, Jordan and Weiss \cite{ng} that we'll denote as the NJW-SC algorithm.

\subsection{An extension to $k$-means: Weighted kernel $k$-means}

A major drawback of the $k$-means algorithm is that it can not separate clusters that are non-linearly separable in input space $\mathcal{S}$ (i.e., if the intersection of the convex hull of the clusters is not empty). One approach for tackling this problem is using the \textit{kernel method}, leading to an algorithm called kernel $k$-means.

\subsubsection{The kernel method}

\begin{figure}[h]%
    \centering
    \subfloat[Concentric circles dataset]{{\includegraphics[width=0.25\textwidth]{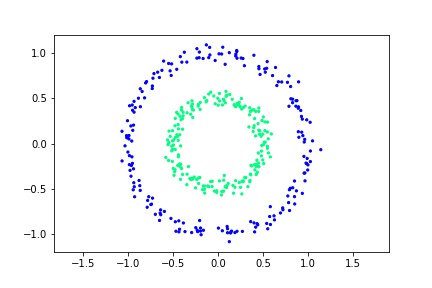} }}%
    \qquad
    \subfloat[Concentric circles dataset in space $\mathcal{F}$]{{\includegraphics[width=0.25\textwidth]{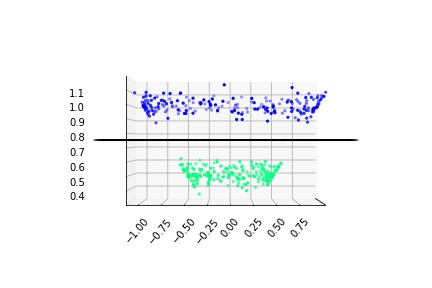} }}%
    \qquad
    \subfloat[Hyperplane separating the concentric circles dataset in $\mathcal{F}$]{{\includegraphics[width=0.25\textwidth]{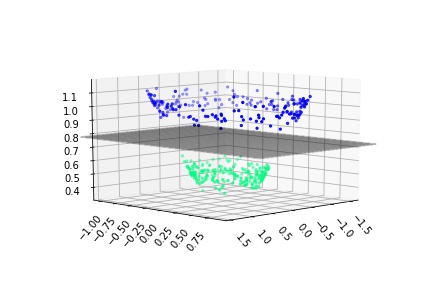} }}%
    \caption{An example of the kernel method}%
    \label{fig:kernelMethod}%
\end{figure}

The kernel method is commonly used in different Machine Learning algorithms. The basic idea behind the kernel method is to map each data point into a different space $\mathcal{F}$ (usually a high dimensional space) via a map $\phi: \mathcal{S} \to \mathcal{F}$, and then use the learning algorithm (in our case, the $k$-means algorithm) in this space $\mathcal{F}$. In our specific case, we would like to have a map $\phi$ such that in the space $\mathcal{F}$ the data points are now linearly separable. Figure \ref{fig:kernelMethod} shows a simple example of the kernel method. The concentric circles dataset shown in figure \ref{fig:kernelMethod}(a) is not linearly separable in the original (two dimensional) space $\mathcal{S}$, but using the mapping $\phi(x_i) = [x_i, \norm{x_i}]$ (i.e., using the distance from any point $x_i$ to the origin as its third coordinate in the three dimensional space $\mathcal{F}$), now the circles are linearly separable in $\mathcal{F}$, as shown in figures \ref{fig:kernelMethod}(b) and \ref{fig:kernelMethod}(c).

One of the key ideas behind the kernel method is to realize that most distance functions (like the Euclidean distance) can be expressed in terms of dot products. Therefore, even if now the data points lies in a possibly high dimensional space, if we can compute the dot product between any pair of points in an efficient manner, then we can compute the distance between them also in an efficient way. This is exactly what a \textit{kernel function} is: a function $\kappa: \mathcal{S} \times \mathcal{S} \to \mathbb{R}$ which computes the dot product between a couple of (mapped) points $\phi(x_i)$ and $\phi(x_j)$ without explicitly computing the mappings $\phi(x_i)$ and $\phi(x_j)$, i.e., $\kappa(x_i, x_j) = \phi^T(x_i)\phi(x_j)$. Popular kernel functions are the polynomial kernel $\kappa_P(x_i, x_j) = (x^T_ix_j + c)^b$, $c,b \in \mathbb{R}$ or the Gaussian kernel $\kappa_G(x_i, x_j) = \exp{\Big(-\frac{\norm{x_i - x_j}^2}{2\sigma^2}\Big)}$, $\sigma \in \mathbb{R}$.

\subsubsection{Weighted kernel $k$-means}

Another extension to the $k$-means algorithm is to associate each data point $x_i$ with a weight $w_i \in \mathbb{R}$. Depending on the context, weight $w_i$ usually express how important point $x_i$ is. Adding both the kernel method and the weights to the optimization formulation of the $k$-means problem \eqref{eq:kmm_opt_prob} results in the following optimization problem:

\begin{equation}
\begin{aligned} \label{eq:kmm_gen_opt_prob}
\min_{r_{i,c}} \quad & \sum^k_{c = 1}\sum_{i \in \Gamma_c}w_i\norm{\phi(x_i) - m_c}^2 \\
\textrm{s.t.} \quad & r_{i,c} \in \{0, 1\} \qquad (\forall i = 1, 2, ... n; c = 1, 2, ... k) \\
\end{aligned}
\end{equation}

Note that weights $w_i$ are not optimization variables, they are known constants, an input to the optimization problem. Also, observe that we have used $m_c$ to denote the weighted mean of the points in cluster $\Gamma_c$, but in the space $\mathcal{F}$:

\begin{equation}
\begin{aligned} \label{eq:kmm_kernel_mu}
m_c = \frac{\sum_{i \in \Gamma_c}w_i\phi(x_i)}{s_c}
\end{aligned}
\end{equation}

where $s_c = \sum_{i \in \Gamma_c}w_i$ is the total weight of cluster $\Gamma_c$. Observe that $m_c$ is the point that minimizes

\begin{equation*}
\sum_{i \in \Gamma_c}w_i\norm{\phi(x_i) - z}^2
\end{equation*}

w.r.t. $z$, i.e., $m_c$ is the "best" cluster representative for cluster $\Gamma_c$ in space $\mathcal{F}$. We can rewrite the objective function of problem \eqref{eq:kmm_gen_opt_prob} as follows

\begin{align*}
\sum^k_{c = 1}\sum_{i \in \Gamma_c}w_i\norm{\phi(x_i) - m_c}^2 &= \sum^k_{c = 1}\sum_{i \in \Gamma_c}w_i\big(\phi(x_i) - m_c\big)^T\big(\phi(x_i) - m_c\big) \\
&= \sum^k_{c = 1}\sum_{i \in \Gamma_c}w_i\big(\phi^T(x_i)\phi(x_i) - 2\phi^T(x_i)m_c + m^T_cm_c\big) \\
&= \sum^k_{c = 1}\Big(\sum_{i \in \Gamma_c}w_i\phi^T(x_i)\phi(x_i) - 2\sum_{i \in \Gamma_c}w_i\phi^T(x_i)m_c + \sum_{i \in \Gamma_c}w_im^T_cm_c\Big) \\
&= \sum^k_{c = 1}\Big(\sum_{i \in \Gamma_c}w_i\phi^T(x_i)\phi(x_i) - 2s_cm^T_cm_c + s_cm^T_cm_c\Big) \\
&= \sum^k_{c = 1}\Big(\sum_{i \in \Gamma_c}w_i\phi^T(x_i)\phi(x_i) - s_cm^T_cm_c\Big) \\
&= \sum^k_{c = 1}\sum_{i \in \Gamma_c}w_i\phi^T(x_i)\phi(x_i) - \sum^k_{c = 1}s_cm^T_cm_c
\end{align*}

where fourth equality is due equation \eqref{eq:kmm_kernel_mu}. Also, note that the first term in the last equation does not depend on the assignments $r_{i,c}$ or the cluster centers $m_c$, so we can remove it from the objective function, obtaining the following objective function

\begin{equation} \label{eq:km_kernel_obj}
- \sum^k_{c = 1}s_cm^T_cm_c = - \sum^k_{c = 1}\frac{\sum_{i,j \in \Gamma_c}w_iw_j\phi^T(x_i)\phi(x_j)}{s_c}
\end{equation}

From the form of this equation, we can see that the quantity to minimize (across all the clusters) is the negative sum of the weighted dot product (in space $\mathcal{F}$) between all pair of points in the same cluster, normalized by the total weight of the cluster. In other words, and interpreting the dot product as a similarity measure (in terms of the direction of the data points), we are trying to \underline{maximize the normalized similarity among points in the same cluster}, where the normalization term is the weight of the cluster, and the similarity function between two points is defined by the kernel $\kappa()$.

Letting $W \in \mathbb{R}^{n \times n}$ be the diagonal matrix with all the weights $w_i$ in the diagonal, $W_c \in \mathbb{R}^{|\Gamma_c| \times |\Gamma_c|}$ (where $|\Gamma_c|$ denotes the number of elements in cluster $\Gamma_c$) be the diagonal matrix of the weights in cluster $\Gamma_c$, $\Phi \in \mathbb{R}^{dim(\mathcal{F}) \times n}$ (where $dim(\mathcal{F})$ denotes the dimension of space $\mathcal{F}$) be the matrix formed by horizontally concatenating the points $\phi(x_i)$ (i.e., $\Phi = [\phi(x_1), \phi(x_2), ..., \phi(x_n)]$), $\Phi_c \in \mathbb{R}^{dim(\mathcal{F}) \times |\Gamma_c|}$ the matrix formed by horizontally concatenating the points $\phi(x_i)$ belonging to cluster $\Gamma_c$ and $e_c$ the vector of ones of size $|\Gamma_c|$, we can rewrite $m_c$ as

\begin{equation}
\begin{aligned} \label{eq:kmm_kernel_mu_2}
m_c = \frac{\Phi_cW_ce_c}{s_c}
\end{aligned}
\end{equation}

Using this equation, we can rewrite equation \eqref{eq:km_kernel_obj} as

\begin{align*}
- \sum^k_{c = 1}s_cm^T_cm_c &= - \sum^k_{c = 1}\frac{e^T_cW_c\Phi^T_c\Phi_cW_ce_c}{s_c} \\
&= - \sum^k_{c = 1}\frac{e^T_cW_c\Phi^T_c}{\sqrt{s_c}}\frac{\Phi_cW_ce_c}{\sqrt{s_c}} \\
&= -Tr(Y^TW^{\frac{1}{2}}\Phi^T\Phi W^{\frac{1}{2}}Y) \\
&= -Tr(Y^TW^{\frac{1}{2}}KW^{\frac{1}{2}}Y)
\end{align*}

where $Tr(A)$ denotes the trace of matrix $A$, $K=\Phi^T\Phi$ is the kernel matrix of the data, i.e., $K_{ij} = \kappa(x_i,x_j)$, and

\begin{equation} \label{eq:y_matrix}
Y = \begin{bmatrix}
\frac{W^{\frac{1}{2}}_1e_1}{\sqrt{s_1}} & 0 & 0 & 0 \\
0 & \frac{W^{\frac{1}{2}}_2e_2}{\sqrt{s_2}} & 0 & 0 \\
0 & 0 & ... & 0 \\
0 & 0 & 0 & \frac{W^{\frac{1}{2}}_ke_k}{\sqrt{s_k}} \\
\end{bmatrix}
\end{equation}

is the assignment matrix, where all the zeros denotes zero matrices of appropiate size. Note that $Y$ is an $n \times k$ orthonormal matrix, i.e., $Y^TY = I$. We can then re-cast the optimization problem \eqref{eq:kmm_gen_opt_prob} as

\begin{equation}
\begin{aligned} \label{eq:eig_kmm_opt_prob}
\max_{Y \in \nu_Y^{n \times k}} \quad & Tr(Y^TW^{\frac{1}{2}}KW^{\frac{1}{2}}Y) \\
\textrm{s.t.} \quad & Y^TY = I
\end{aligned}
\end{equation}

where $\nu_Y = \{0\} \bigcup \left\{\sqrt{\frac{w_i}{s_c}}\right\}, i \in \Gamma_c$. We will denote problem \eqref{eq:eig_kmm_opt_prob} as the matrix version of the weighted kernel $k$-means problem. There are two important things to observe from problem \eqref{eq:eig_kmm_opt_prob}. The first one is that, if we set all the weights $w_i$ equal to one and set the function $\phi()$ to be the identity function, then problem \eqref{eq:kmm_gen_opt_prob} turns into problem \eqref{eq:kmm_opt_prob}, the vanilla $k$-means problem, and the corresponding matrix version of this problem turns into the following optimization problem

\begin{equation}
\begin{aligned} \label{eq:eig_vanilla_kmm_opt_prob}
\max_{Y \in \nu_Y^{n \times k}} \quad & Tr(Y^TGY) \\
\textrm{s.t.} \quad & Y^TY = I
\end{aligned}
\end{equation}

where $\nu_Y = \{0\} \bigcup \left\{\sqrt{\frac{1}{s_c}}\right\}, i \in \Gamma_c$, and $G$ is the Gram matrix of the data points, i.e., $G_{ij} = x^T_ix_j$. We will denote problem \eqref{eq:eig_vanilla_kmm_opt_prob} as the matrix version of the vanilla $k$-means problem.

The second thing is that, just like the $k$-means algorithm can find a local minima for the vanilla $k$-means problem, we can use a modified version of the $k$-means algorithm to find a local minima for the weighted kernel $k$-means problem. This is shown in algorithm \ref{algo:wk_kmeans}. There is something important to highlight with this algorithm. Following it naively will get you into a situation where you need the mapped points $\phi(x_i)$ in both steps \textbf{E} and \textbf{M}. However, as we mentioned previously, the whole idea of the kernel method is to avoid the explicit computation of $\phi(x_i)$, and instead use only the dot products $\phi^T(x_i)\phi(x_j) = \kappa(x_i,x_j)$. So, for step \textbf{E}, when computing the squared distance $||\phi(x_i) - m_j||^2$, it is implied that we need to use the representation of the mean $m_j$ using equation \eqref{eq:kmm_kernel_mu}, substitute it in the equation of the squared distance, re-write this equation as a dot product, and then use the distributive property of the dot product to express $||\phi(x_i) - m_j||^2$ as a sum of weighed dot products $\phi^T(x_i)\phi(x_k)$ that we can compute using the kernel function $\kappa()$. Finally, observe that we don't need to compute the value of $m_c$ in the \textbf{M} step, because, as mentioned previously, it can be only implicitly computed via $\kappa()$ (we can't explicitly compute it, because we don't have access to the mapped points $\phi(x_i)$). This is the reason why, in contrast to the vanilla $k$-means algorithm, the weighted kernel $k$-means algorithm only returns the assignments $r_{i,c}$.

\begin{algorithm}
\DontPrintSemicolon 
\KwIn{Dataset $X$, number of clusters $k$, weights $w_i$, kernel function $\kappa()$ and initial values for every $m_c$}
\KwOut{Learned values for $r_{i,c}$}
\While{Convergence criterion is not satisfied}{
\textbf{E step}. Using the current values of $m_c$, assign every point to its closest cluster: \\

\begin{equation*}
r_{i,c} \gets
\begin{cases}
1 & \text{if $c = argmin_j||\phi(x_i) - m_j||^2$}\\
0 & \text{otherwise}
\end{cases}       
\end{equation*}

\textbf{M step}. Re-estimate the cluster means using the current assignations $r_{i,c}$:

\begin{equation*}
m_c \gets \frac{\sum_{i \in \Gamma_c}w_i\phi(x_i)}{s_c}
\end{equation*}
}
\Return{$\boldsymbol{r}$}\;
\caption{Weighted kernel $k$-means}
\label{algo:wk_kmeans}
\end{algorithm}

\subsection{An spectral solution for the Weighted kernel $k$-means problem}

The spectral decomposition of matrix $W^{\frac{1}{2}}KW^{\frac{1}{2}}$ can provide a different solution for the weighted kernel $k$-means problem. A standard result in linear algebra \cite{golub} provides a global solution to a relaxed version of problem \eqref{eq:eig_kmm_opt_prob}. By allowing $Y$ to be an arbitrary orthonormal matrix, we can obtain an optimal Y by taking the top $k$ eigenvectors (i.e., the eigenvectors associated to the $k$ largest eigenvalues) of the matrix $W^{\frac{1}{2}}KW^{\frac{1}{2}}$. Each row of the resulting matrix $Y$ is then interpreted as an "embedding" version of the original data point in a lower dimensional space with dimension equal to $k$. The typical way to proceed is to compute a discrete partition of the embedded data points, usually using the vanilla $k$-means algorithm with the embedded points.

\subsubsection{The NJW-SC algorithm}

The NJW-SC (Ng-Jordan-Weiss Spectral Clustering) algorithm, shown in algorithm \ref{algo:SC}, considers the weighted kernel $k$-means problem, using the Gaussian kernel 

\begin{equation}
K_{ij} = \exp{\Big(-\frac{\norm{x_i - x_j}^2}{2\sigma^2}\Big)}
\end{equation}

and setting 

\begin{equation} \label{eq: njw_w}
W^{-1} = diag(K1_n)
\end{equation}

where $diag(v)$ is a square diagonal matrix with the elements of vector $v$ on the diagonal, and $1_n$ denotes the vector of ones of size $n$. Then, it proceeds to compute the spectral solution mentioned in the last section: it computes the top $k$ eigenvectors of the matrix $W^{\frac{1}{2}}KW^{\frac{1}{2}}$, and use the embedded data points to compute a discrete partition of the data.

\subsubsection{NJW-SC complexity}

Using a kernel gives the NJW-SC algorithm more flexibility than the vanilla $k$-means algorithm. However, the cost to pay for this flexibility is the time complexity of the algorithm. In the first step, the \textbf{Kernel matrix step}, we need to compute the kernel function for $\frac{n(n+1)}{2}$ pairs of points. Therefore, the complexity of this step is $O(n^2d)$, where $O(d)$ is the complexity of computing the Gaussian kernel $\kappa_G()$. In the second step, the \textbf{Weights matrix step}, we need to compute the matrix of weights $W$ using equation \eqref{eq: njw_w}, which implies a matrix-vector multiplication, with complexity $O(n^2)$. The third step, the \textbf{Eigendecomposition step}, consist of two sub-steps. The first one is the matrix multiplication $W^{\frac{1}{2}}KW^{\frac{1}{2}}$, which can take up to $O(n^3)$ operations, but given that $W$ is a diagonal matrix, this sub-step has a complexity of $O(n^2)$. The second sub-step, the computation of the top $k$ eigenvectors, can take up to $O(n^3)$ operations if computed naively (i.e., computing the full eigendecomposition). However, as we are interested only in the top $k$ eigenvectors, this sub-step can be computed with more efficient techniques, such as the \textit{power method}, that scale like $O(n^2k)$ \cite{bishop}. The complexity of the last step, the \textbf{Partition step}, depends on the method used to create the partition. If vanilla $k$-means is used for this purpose, the complexity of this step is $O(nk^2T)$. As we can see, and assuming $k << d,n$ (which typically is true in practice), the complexity of the algorithm is $O(n^2d)$, due the computation of the kernel matrix in the first step. This complexity and other important properties of the NJW-SC algorithm are summarized in table \ref{tab:SC}.



\begin{algorithm}
\DontPrintSemicolon 
\KwIn{Dataset $X$, number of clusters $k$}
\KwOut{Learned values for $r_{i,c}$}
\textbf{Kernel matrix step}. Compute kernel matrix $K$ for dataset $X$: \\

\begin{equation*}
K_{ij} \gets \exp{\Big(-\frac{\norm{x_i - x_j}^2}{2\sigma^2}\Big)};
\end{equation*}

\textbf{Weights matrix step}. Compute matrix of weights $W$: \\

\begin{equation*}
W \gets diag^{-1}(K1_n);
\end{equation*}

\textbf{Eigendecomposition step}. Compute top $k$ eigenvectors: \\

\begin{equation*}
Y \gets topEig(W^{\frac{1}{2}}KW^{\frac{1}{2}},k);
\end{equation*}

\textbf{Partition step}. Using the embedded points in $Y$, compute a partition $\boldsymbol{r}$ of the data \\
\Return{$\boldsymbol{r}$}\;
\caption{NJW-SC}
\label{algo:SC}
\end{algorithm}

\begin{table}[ht]
\centering
\caption{NJW-SC algorithm}
\begin{tabular}[t]{lc}
\hline
&SC\\
\hline
Number of hyperparameters & 2 \\
Hyperparameters & number of clusters $k$, width of Gaussian kernel $\sigma$   \\
Time complexity & $O(n^2d)$ \\
Outlier detection & No \\
Type of assignment & hard \\
distance/similarity measure & Gaussian kernel $\kappa_G()$ (See Spectral Clustering section for details) \\
\hline
\end{tabular}
\label{tab:SC}
\end{table}

\subsection{From graphs to Spectral Clustering}

In \cite{dhillon}, Dhillon, Guan and Kulis present a different formulation of the NJW-SC algorithm using a graph cut approach. This formulation will be useful for next section.

\subsubsection{Normalized graph cuts}

Given a graph $G=(\mathcal{V},\mathcal{E},A)$, where $\mathcal{V}$ is a set of $n$ nodes (or vertices), $\mathcal{E}$ is the set of edges connecting nodes, and $A \in \mathbb{R}^{n \times n}$ is a non-negative and symmetric \textit{edge similarity matrix}, i.e., $A_{ij}$ is a (non-negative) weight expressing the similarity between node $i$ and node $j$, the $k$-way \textit{normalized cut problem} is defined as follows. Suppose $\mathcal{A},\mathcal{B} \subseteq \mathcal{V}$, and define:

\begin{equation}
links(\mathcal{A},\mathcal{B}) = \sum_{i \in \mathcal{A},j \in \mathcal{B}} A_{ij}
\end{equation}

i.e., $links(\mathcal{A},\mathcal{B})$ is the sum of the weights of the edges that cross from subset $\mathcal{A}$ to subset $\mathcal{B}$, and:

\begin{equation}
normlinks(\mathcal{A},\mathcal{B}) = \frac{links(\mathcal{A},\mathcal{B})}{links(\mathcal{A},\mathcal{V})}
\end{equation}

is the normalized link ratio of $\mathcal{A},\mathcal{B}$. The $k$-way normalized cut problem is to minimize the links that "escape" a cluster $\mathcal{V}_c \subseteq \mathcal{V}$ relative to the total "weight" of the cluster, where the set of clusters $\mathcal{V}_c$ is a partition of $\mathcal{V}$:

\begin{equation}
\begin{aligned} \label{eq:graph_cut_opt_prob}
\min_{\mathcal{V}_c} \quad & \frac{1}{k}\sum^k_{c = 1}normlinks(\mathcal{V}_c,\mathcal{V} \setminus \mathcal{V}_c) \\
\end{aligned}
\end{equation}

where $\mathcal{V} \setminus \mathcal{V}_c$ denotes the set of nodes in $\mathcal{V}$ that do not belong to cluster $\mathcal{V}_c$. We can rewrite the objective function of problem \eqref{eq:graph_cut_opt_prob} as follows

\begin{align*}
\frac{1}{k}\sum^k_{c = 1}normlinks(\mathcal{V}_c,\mathcal{V} \setminus \mathcal{V}_c) &= \frac{1}{k}\sum^k_{c = 1}\frac{links(\mathcal{V}_c,\mathcal{V} \setminus \mathcal{V}_c)}{links(\mathcal{V}_c,\mathcal{V})} \\
&= \frac{1}{k}\sum^k_{c = 1}\frac{\sum_{i \in \mathcal{V}_c,j \notin \mathcal{V}_c} A_{ij}}{\sum_{i \in \mathcal{V}_c,j \in \mathcal{V}} A_{ij}} \\
&= \frac{1}{k}\sum^k_{c = 1}\frac{\sum_{i \in \mathcal{V}_c,j \in \mathcal{V}} A_{ij} - \sum_{i,j \in \mathcal{V}_c} A_{ij}}{\sum_{i \in \mathcal{V}_c,j \in \mathcal{V}} A_{ij}} \\
&= \frac{1}{k}\sum^k_{c = 1}(1 - normlinks(\mathcal{V}_c,\mathcal{V}_c)) \\
&= 1 - \frac{1}{k}\sum^k_{c = 1}\frac{\sum_{i,j \in \mathcal{V}_c} A_{ij}}{\sum_{i \in \mathcal{V}_c,j \in \mathcal{V}} A_{ij}} \\
\end{align*}

We can observe that problem \eqref{eq:graph_cut_opt_prob} is equivalent to the following optimization problem

\begin{equation}
\begin{aligned} \label{eq:graph_cut_opt_prob_max}
\max \quad & \sum^k_{c = 1}\frac{\sum_{i,j \in \mathcal{V}_c} A_{ij}}{\sum_{i \in \mathcal{V}_c,j \in \mathcal{V}} A_{ij}} \\
\end{aligned}
\end{equation}

In other words, we are trying to \underline{maximize the normalized similarity among points in the same cluster}, where the normalization term is the "weight" of the cluster, and the similarity between two points is defined by the similarity matrix $A$. Letting $D \in \mathbb{R}^{n \times n}$ be the diagonal matrix with node weights $d_i$ in the diagonal, where a node weight $d_i$ is defined as the sum of weights $A_{ij}$ for edges adjacent to node $i$, i.e., $d_i = \sum^n_{j=1}A{ij}$, and $U \in \{0,1\}^{n \times k}$ be the assignment matrix where $U_{ic} = 1$ if node $i$ belongs to cluster $\mathcal{V}_c$ and $0$ otherwise, then we can rewrite the objective function of problem \eqref{eq:graph_cut_opt_prob_max} as 

\begin{align*}
\sum^k_{c = 1}\frac{\sum_{i,j \in \mathcal{V}_c} A_{ij}}{\sum_{i \in \mathcal{V}_c,j \in \mathcal{V}} A_{ij}} &= \sum^k_{c = 1}\frac{U^T_cAU_c}{U^T_cDU_c} \\
&= Tr\Big((U^TDU)^{-\frac{1}{2}}U^TAU(U^TDU)^{-\frac{1}{2}}\Big) \\
\end{align*}

where $U_c$ is the $c$-th column of matrix $U$, corresponding to cluster $\mathcal{V}_c$, and, as can be verified,

\begin{equation} \label{eq:udu_matrix}
(U^TDU)^{-\frac{1}{2}} = \begin{bmatrix}
\frac{1}{\sqrt{U^T_1DU_1}} & 0 & ... & 0 \\
0 & \frac{1}{\sqrt{U^T_2DU_2}} & ... & 0 \\
\vdots & \vdots & \ddots & \vdots \\
0 & 0 & ... & \frac{1}{\sqrt{U^T_kDU_k}} \\
\end{bmatrix}
\end{equation}

and

\begin{equation} \label{eq:uau_matrix}
(U^TAU) = \begin{bmatrix}
U^T_1AU_1 & U^T_1AU_2 & ... & U^T_1AU_k \\
U^T_2AU_1 & U^T_2AU_2 & ... & U^T_2AU_k \\
\vdots & \vdots & \ddots & \vdots \\
U^T_kAU_1 & U^T_kAU_2 & ... & U^T_kAU_k \\
\end{bmatrix}
\end{equation}

Note that, in order to ensure that all points are assigned to exactly one cluster, any row of matrix $U$ must have exactly one entry with value one, and the rest equal to zero, i.e., $U1_k = 1_n$. Also, observe that

\begin{equation} \label{eq:uau_matrix2}
(U^TU) = \begin{bmatrix}
|\mathcal{V}_1| & 0 & ... & 0 \\
0 & |\mathcal{V}_2| & ... & 0 \\
\vdots & \vdots & \ddots & \vdots \\
0 & 0 & ... & |\mathcal{V}_k| \\
\end{bmatrix}
\end{equation}

We will denote this matrix as $I_{U^TU}$. We can then re-cast the optimization problem \eqref{eq:graph_cut_opt_prob_max} as

\begin{equation}
\begin{aligned} \label{eq:graph_cut_opt_prob_max_matrix_1}
\max_{U \in \{0,1\}^{n \times k}} \quad & Tr\Big((U^TDU)^{-\frac{1}{2}}U^TAU(U^TDU)^{-\frac{1}{2}}\Big) \\
\textrm{s.t.} \quad & U^TU = I_{U^TU} \\
\quad & U1_k = 1_n \\
\end{aligned}
\end{equation}

 Finally, letting $\Tilde{Z} = D^{\frac{1}{2}}U(U^TDU)^{-\frac{1}{2}}$ results in the following optimization problem

\begin{equation}
\begin{aligned} \label{eq:graph_cut_opt_prob_max_matrix_2}
\max_{\Tilde{Z} \in \nu_{\Tilde{Z}}^{n \times k}} \quad & Tr\Big(\Tilde{Z}^TD^{-\frac{1}{2}}AD^{-\frac{1}{2}}\Tilde{Z}\Big) \\
\textrm{s.t.} \quad & \Tilde{Z}^T\Tilde{Z} = I \\
\end{aligned}
\end{equation}

where $\nu_{\Tilde{Z}} = \{0\} \bigcup \left\{\sqrt{\frac{d_i}{U^T_cDU_c}}\right\}, i \in \Gamma_c$. We will denote problem \eqref{eq:graph_cut_opt_prob_max_matrix_2} as the matrix version of the $k$-way normalized cut problem.

Comparing  problem \eqref{eq:graph_cut_opt_prob_max_matrix_2} with problem \eqref{eq:eig_kmm_opt_prob}, we can observe that they are equivalent problems: matrices $\Tilde{Z}$ and $Y$ are analogous, both orthonormal assignment matrices with elements equal to zero or to the square root of a normalized weight corresponding to a node ($d_i$) or data point ($w_i$). Matrices $A$ and $K$ are both similarity matrices: matrix $A$ in a explicit way and matrix $K$ as a kernel matrix. Finally, and more specifically, the NJW-SC algorithm sets the matrix $W = diag^{-1}(K1_n)$, and, we can rewrite matrix $D$ as

\begin{equation}
D = diag(A1_n)    
\end{equation}

\begin{figure}[h]%
    \centering
    \subfloat[Banana dataset]{{\includegraphics[width=0.45\textwidth]{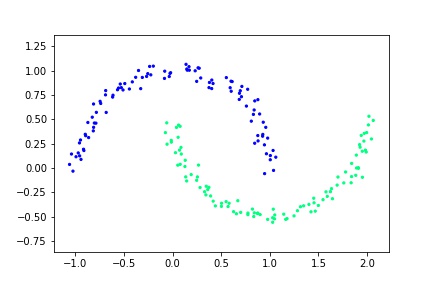} }}%
    \qquad
    \subfloat[Graph created from the banana dataset$^1$]{\includegraphics[width=0.45\textwidth]{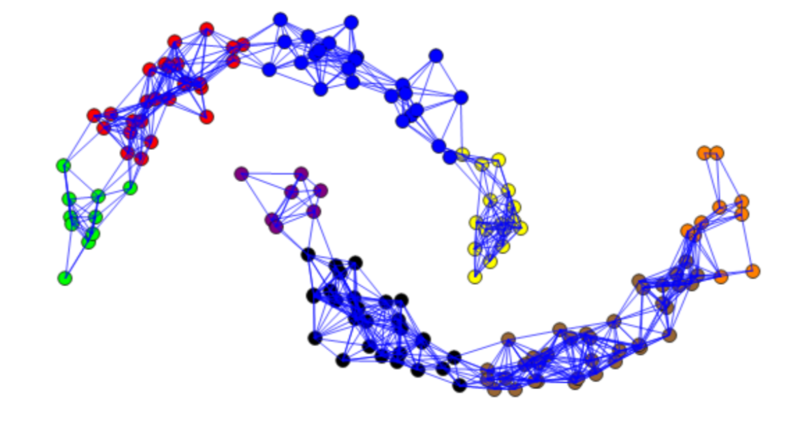} }%
    \caption{Banana dataset and the corresponding graph created by a SC algorithm}%
    \label{fig:bananaSC}%
\end{figure}

Therefore, we can think of the NJW-SC algorithm as an algorithm that reduces a data clustering problem to a graph normalized cut problem: it takes the data points in $X$ and creates a complete graph $G = (\mathcal{V}, \mathcal{E}, A)$, where every pair of nodes is connected by a unique edge. This is shown in figure \ref{fig:bananaSC}. The graph shown is for illustration purpose only: as mentioned before, the NJW-SC algorithm creates a fully connected graph. 

\footnotetext[1]{Image taken from \cite{ucdenver}}

For every data point in $X$, the NJW-SC algorithm adds a corresponding node in $\mathcal{V}$, and it sets the similarity matrix $A$ equal to the kernel matrix $K$ of the data. Finally, it solves the matrix version of the $k$-way normalized cut problem \eqref{eq:graph_cut_opt_prob_max_matrix_2} computing the top $k$ eigenvectors of matrix $D^{-\frac{1}{2}}AD^{-\frac{1}{2}}$ and using them to create a partition of the nodes in $\mathcal{V}$. This is shown in algorithm \ref{algo:NJW-SC-Graph}.

This interesting connection between graph theory and clustering algorithms will allow us to find a relation between the NJW-SC and the DBSCAN algorithms.

\begin{algorithm}
\DontPrintSemicolon 
\KwIn{Dataset $X$, number of clusters $k$}
\KwOut{$k$ clusters in $X$}

\textbf{Graph step}. Create a complete graph $G = (\mathcal{V}, \mathcal{E}, A)$ such that, for every point in $X$, there is a corresponding node in $\mathcal{V}$, and for any pair of nodes in $\mathcal{V}$ there is an edge in $\mathcal{E}$. Set the similarity matrix $A$ as:

\begin{equation*}
A_{ij} \gets \exp{\Big(-\frac{\norm{x_i - x_j}^2}{2\sigma^2}\Big)};
\end{equation*}

\textbf{Weights matrix step}. Compute matrix of weights $D$: \\

\begin{equation*}
D \gets diag(A1_n);
\end{equation*}

\textbf{Eigendecomposition step}. Compute top $k$ eigenvectors: \\

\begin{equation*}
\Tilde{Z} \gets topEig(D^{-\frac{1}{2}}AD^{-\frac{1}{2}},k);
\end{equation*}

\textbf{Partition step}. Using the embedded points in $\Tilde{Z}$, compute a partition $\mathcal{V}_c$ of the nodes in $\mathcal{V}$ \\
\Return{$\mathcal{V}_c$, $\forall c \in \{1,2,...,k\}$}\;
\caption{NJW-SC, Graph version}
\label{algo:NJW-SC-Graph}
\end{algorithm}

\section{DBSCAN}

The main difference between DBSCAN and any of the previous algorithms explored is that DBSCAN is a density based algorithm: it uses two parameters $\epsilon$ and $minPts$ to compute the local density of every data point $x_i$ and uses this information to create a set of cluster from the data. Observe that the word density is being used loosely: in reality, what DBSCAN computes is the number of neighbors inside its $\epsilon$-neighborhood $N_\epsilon(x_i) = \{x_j \in X | \norm{x_i - x_j} \leq \epsilon\}$. If the number of neighbors in $N_\epsilon(x_i)$ is equal or larger than $minPts$ (which stands for \textit{minimum number of points}), then we label $x_i$ as a \textit{core point} (or \textit{dense point}). 

\subsection{SC but faster (and sparser)}

Three of the main disadvantages of the NJW-SC algorithm are:

\begin{itemize}
    \item It needs to find the top $k$ eigenvectors of a $n \times n$ matrix. Despite not being the term that dominates the complexity of the algorithm, this step adds $O(n^2k)$ computations to the algorithm
    \item Choice of $k$. Without any prior knowledge, the choice of $k$ can be difficult
    \item No outlier detection. The algorithm can't detect outliers. It will assign any outlier to some cluster, or it will create a size-one cluster with the outlier
\end{itemize}

We will see how using a different kernel, and introducing a "filtering" step into the NJW-SC algoritm we can create an algorithm with none of the mentioned disadvantages.

\subsubsection{A discontinuous kernel}

The Gaussian kernel, used in the NJW-SC algorithm, is a continuous kernel: the function $\kappa_G()$ is continuous w.r.t. its argument $\norm{x_i - x_j}^2$. Using this kernel have some advantages: all the similarity values lies in the $[0,1]$ interval, and the kernel returns $1$ if and only if $x_i = x_j$. However, it also has some disadvantages. One of them is that, no matter how far two points are, the similarity between them is never $0$. Zero is only achieved in the limit $\norm{x_i - x_j}^2 \rightarrow \infty$. Using a different kernel we might overcome this disadvantage. Let $\kappa_H()$ be the Heaviside kernel (based on the Heaviside function), defined as:

\begin{equation}
\kappa_H(x_i,x_j) =
\begin{cases}
1 & \text{if $\epsilon - \norm{x_i - x_j} \geq 0$}\\
0 & \text{otherwise}
\end{cases}
\end{equation}

\begin{figure}[h]
    \centering
    \includegraphics[width=0.45\textwidth]{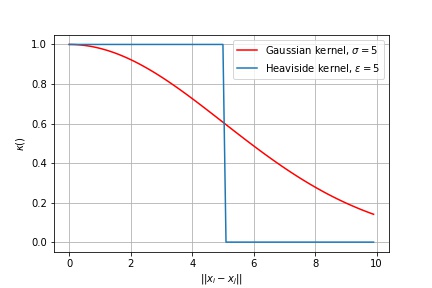}
    \caption{Gaussian and Heaviside kernels}
    \label{fig:kernels}
\end{figure}

From this form, we can see that the Heaviside kernel is discontinuous, as shown in figure \ref{fig:kernels}. This kernel assigns a similarity value of one to any two data points if the distance between them is less or equal than $\epsilon$, and zero otherwise. Using this discontinuous kernel will allows to modify the NJW-SC algorithm in order to create a faster algorithm.

\subsubsection{Core points}

Modifying algorithm \ref{algo:NJW-SC-Graph} using the Heaviside kernel instead of the Gaussian kernel is the first modification that we will do to the NJW-SC algorithm. The second modification will be the following. After the \textbf{Weights matrix step}, we introduce a filtering step: we will filter the nodes in $G$ using their weights $d_i = D_{ii}$ (in this case, due the use of the Heaviside kernel, the weight $d_i$ is equal to the degree of its corresponding node, plus one). This filtering process will be simple:

\begin{itemize}
    \item If $d_i = 1$ (i.e., if node $i$ is an isolated node), we will label the corresponding node (and therefore, the corresponding data point $x_i$) as an \textit{outlier}, removing it from the graph $G$, with all its adjacent edges, and we will remove the corresponding row and column from the matrix $A$
    \item If $1 < d_i \leq minPts$, the corresponding node (and therefore, the corresponding data point $x_i$) will be labeled as unprocessed, removing it from the graph $G$, with all its adjacent edges, and we will remove the corresponding row and column from the matrix $A$
    \item Finally, if $d_i > minPts$, we will label the corresponding node (and therefore, the corresponding data point $x_i$) as a \textit{core node}, preserving it in the graph $G$, with all its adjacent edges, and preserving the corresponding row and column in the matrix $A$
\end{itemize}


We have three different type of data points after the filtering process: (current) outliers, unprocessed points and core points. We will see that, once the algorithm has finished, any unprocessed point will be either a part of a cluster or an outlier. Also, observe that we have reduced the size of matrices $A$ and $D$ from $n \times n$ to $n_{core} \times n_{core}$, where $n_{core}$ is the number of core points found in the dataset. From this fact, we can see that we had improved the complexity of the next step (the \textbf{Eigendecomposition step}) from $O(n^2k)$ to $O(n_{core}^2k)$. Also, observe that, although we haven't done it explicitly, if two points have a similarity value of zero, the corresponding edge will be removed (or more precisely, not taken into account) from graph $G$. Therefore, we can think of (post-filtered) matrix $A$ as the sum of the adjacency matrix $A_G$ of (post-filtered) graph $G$ plus the identity matrix of appropiate size

\begin{equation} \label{eq:graphA}
A = A_G + I  
\end{equation}

Similarly, we can think of (post-filtered) matrix $D$ as the sum of the degree matrix $D_G$ of (post-filtered) graph $G$ plus the identity matrix of appropiate size

\begin{equation} \label{eq:graphD}
D = D_G + I  
\end{equation}

There is something important to highlight with respect to this last point. Since matrix A has been modified, the post-filtered matrix $D$ needs to be recomputed $D = diag(A1_n)$. The \textit{Laplacian matrix} $\mathcal{L} \in \mathbb{R}^{n_{core} \times n_{core}}$ of (post-filtered) graph $G$ is defined as:

\begin{equation} \label{eq:graphL}
\mathcal{L} = D_G - A_G = D_G - A_G + (I - I) = D - A
\end{equation}

where the last equality is due equations \eqref{eq:graphA} and \eqref{eq:graphD}. Given the Laplacian matrix $\mathcal{L}$ and the degree matrix $D_G$ of a graph $G$, the normalized Laplacian of a matrix is defined as:

\begin{equation} \label{eq:graphNL}
\mathcal{L}_{norm} = D_G^{-\frac{1}{2}}\mathcal{L}D_G^{-\frac{1}{2}}
\end{equation}

A very well known result in Graph Theory is that eigenvectors related to zero eigenvalues of the normalized Laplacian indicate connected components in the graph $G$ \cite{schubert}. The results still holds if we use the matrix $D$ instead of $D_G$ as can be easily verified. Therefore, given an eigenvector $v_0$ related to a zero eigenvalue

\begin{align*}
0 &= \mathcal{L}_{norm}v_0 \\
&= D_G^{-\frac{1}{2}}\mathcal{L}D_G^{-\frac{1}{2}}v_0 \\
&= D^{-\frac{1}{2}}\mathcal{L}D^{-\frac{1}{2}}v_0 \\
&= D^{-\frac{1}{2}}(D - A)D^{-\frac{1}{2}}v_0 \\
&= v_0 - D^{-\frac{1}{2}}AD^{-\frac{1}{2}}v_0 \\
\end{align*}

which implies

\begin{equation} 
D^{-\frac{1}{2}}AD^{-\frac{1}{2}}v_0 = v_0
\end{equation}

i.e., normalized Laplacian's eigenvector $v_0$ is also an eigenvector of the matrix $D^{-\frac{1}{2}}AD^{-\frac{1}{2}}$, in this case with eigenvalue equal to one.

This last observation has an important consequence: the eigenvectors obtained in the next step (\textbf{Eigendecomposition step}) with associated eigenvalue equal to one will be indicator vectors for the connected components in (post-filtered) graph $G$: for any of these eigenvectors, all the non-zero entries corresponds to points belonging to the same cluster. This property will allows to modify the \textbf{Eigendecomposition step} in order to remove the dependency to the parameter $k$.

Just as mentioned in the previous paragraph, in the \textbf{Eigendecomposition step}, instead of using the $k$ top eigenvectors, we will compute all the eigenvectors with eigenvalue equal to one. These eigenvectors will be indicator vectors for the connected components in (post-filtered) $G$, which in turn will be the clusters formed by the core points. Due this, we can remove the last step of the algorithm (the \textbf{Partition step}), because we can read the partition directly from the eigenvectors with eigenvalue equal to one.

Once we have found the clusters in the set of core points, we need to process the unprocessed points. Our modified algorithm does this in a simple way: it will assign each unprocessed point into the same cluster as their closest core point, if this core point is part of its $\epsilon$-neighborhood. Otherwise, it will label it as an outlier.

This modified algorithm is shown in algorithm \ref{algo:DBSCAN_SC}, and is equivalent to the DBSCAN algorithm. However, other versions of the DBSCAN algorithm doesn't compute the connected components of $G$ using the eigendecomposition of matrix $D^{-\frac{1}{2}}AD^{-\frac{1}{2}}$. For instance, in \cite{jang}, they use instead a depth-first search (DFS), a more efficient procedure with complexity $O(n_{core}^2)$. In algorithm \ref{algo:DBSCAN}, we present a different formulation of the DBSCAN algorithm, similar to the version presented in \cite{jang}, and equivalent to the original formulation of the algorithm in \cite{ester}. We prefer to present this version because it is more similar to algorithm \ref{algo:DBSCAN_SC}.

\begin{algorithm}
\DontPrintSemicolon 
\KwIn{Dataset $X$, radius $\epsilon$, and minimum number of points $minPts$}
\KwOut{Dense-connected clusters and outliers in $X$}

\textbf{Graph step}. Create a graph $G = (\mathcal{V}, \mathcal{E}, A)$ such that, for every point in $X$, there is a corresponding node in $\mathcal{V}$, and for any pair of nodes in $\mathcal{V}$ there is an edge in $\mathcal{E}$. Set the similarity matrix $A$ as:

\begin{equation*}
A_{ij} \gets 
\begin{cases}
1 & \text{if $\epsilon \geq \norm{x_i - x_j}$}\\
0 & \text{otherwise}
\end{cases};
\end{equation*}

\textbf{Weights matrix step}. Compute matrix of weights $D$: \\

\begin{equation*}
D \gets diag(A1_n);
\end{equation*}

\textbf{Core points identification step}. Identify the core points: \\

\begin{equation*}
label_i \gets
\begin{cases}
core & \text{if $d_i > minPts$} \\
unprocessed & \text{if $1 < d_i \leq minPts$} \\
outlier & \text{otherwise} \\
\end{cases}       
\end{equation*}

If $label_i$ is equal to $unprocessed$ or $outlier$, remove node $i$ from graph $G$, with all its adjacent edges.Make the corresponding changes to matrices $A$ and $D$

\textbf{Eigendecomposition step}. Compute eigenvectors with eigenvalue equal to one: \\

\begin{equation*}
\Tilde{Z} \gets EigOne(D^{-\frac{1}{2}}AD^{-\frac{1}{2}});
\end{equation*}

The corresponding clusters can be read from $\Tilde{Z}$

\textbf{Outliers identification step}. Assign each $unprocessed$ point in $X$ to the same cluster as their closest core point, if this core point is part of its $\epsilon$-neighborhood. Otherwise, label it as an outlier: \\ 

\begin{equation*}
label_i \gets
\begin{cases}
core & \text{if closest core point $\in N_\epsilon(x_i)$} \\
outlier & \text{otherwise}
\end{cases}       
\end{equation*} \\

\Return{Connected components and outliers in $G$}\;
\caption{DBSCAN, Spectral version}
\label{algo:DBSCAN_SC}
\end{algorithm}

\subsubsection{DBSCAN complexity}

Identifying the core points in $X$ takes up to $O(n^2d)$ operations, because we need to compute the distance between any pair of points in the dataset. Constructing the graph in the \textbf{Graph step} takes $O(n_{core}^2)$ operations, because that is the maximum possible number of edges. Computing the connected components of $G$ using DFS instead of the eigendecomposition reduces the complexity of the \textbf{Connected components step} from $O(n_{core}^2k)$ to $O(n_{core}^2)$. Finally, and assuming that we have stored in memory the distances between any pair of points in $X$, the \textbf{Outliers identification step} takes up to $O(n_{core}(n - n_{core}))$ operations, because we need to find the closest core point for every unprocessed point. It is evident that the \textbf{Core points identification step} is the most computational intensive step in algorithm \ref{algo:DBSCAN}. Therefore, the complexity of DBSCAN is $O(n^2d)$. This complexity and other important properties of DBSCAN are summarized in table \ref{tab:DBSCAN}.

\begin{table}[ht]
\centering
\caption{DBSCAN algorithm}
\begin{tabular}[t]{lc}
\hline
&DBSCAN\\
\hline
Number of hyperparameters & 2 \\
Hyperparameters & radius $\epsilon$, minimum number of points $minPts$  \\
Time complexity & $O(n^2d)$ \\
Outlier detection & Yes \\
Type of assignment & hard \\
distance/similarity measure & Heaviside kernel $\kappa_H()$ (See DBSCAN section for details) \\
\hline
\end{tabular}
\label{tab:DBSCAN}
\end{table}

\newpage


\begin{algorithm}
\DontPrintSemicolon 
\KwIn{Dataset $X$, radius $\epsilon$, and minimum number of points $minPts$}
\KwOut{Dense-connected clusters and outliers in $X$}

\textbf{Core points identification step}. For every $x_i$ in $X$, find the points in its $\epsilon$-neighborhood $N_\epsilon(x_i)$, and identify the core points with more than $minPts$ neighbors: \\ 

\begin{equation*}
label_i \gets
\begin{cases}
core & \text{if $|N_\epsilon(x_i)| \geq minPts$}\\
unprocessed & \text{otherwise}
\end{cases}       
\end{equation*}

\textbf{Graph step}. Create a graph $G = (\mathcal{V}, \mathcal{E})$ such that, for every core point in $X$, there is a corresponding "core"node in $\mathcal{V}$, and add an edge to $\mathcal{E}$ between this node and all "core" nodes in its $\epsilon$-neighborhood $N_\epsilon(x_i)$ \\

\textbf{Connected components step}. Find the connected components of graph $G$ \\

\textbf{Outliers identification step}. Assign each non-core point in $X$ to the same cluster as their closest core point, if this core point is an $\epsilon$-neighbor. Otherwise, label it as an outlier: \\ 

\begin{equation*}
label_i \gets
\begin{cases}
core & \text{if closest core point $\in N_\epsilon(x_i)$} \\
outlier & \text{otherwise}
\end{cases}       
\end{equation*}

\Return{Connected components and outliers in $G$}\;
\caption{DBSCAN, Graph version}
\label{algo:DBSCAN}
\end{algorithm}

\subsection{Another way to find core points: climbing the hill}

\begin{figure}[h]%
    \centering
    \subfloat[Data points in two dimensions]{{\includegraphics[width=0.25\textwidth]{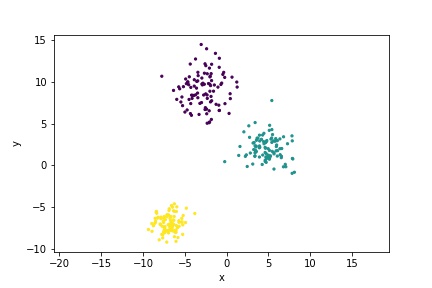} }}%
    \qquad
    \subfloat[Contour plot]{{\includegraphics[width=0.25\textwidth, height=80pt]{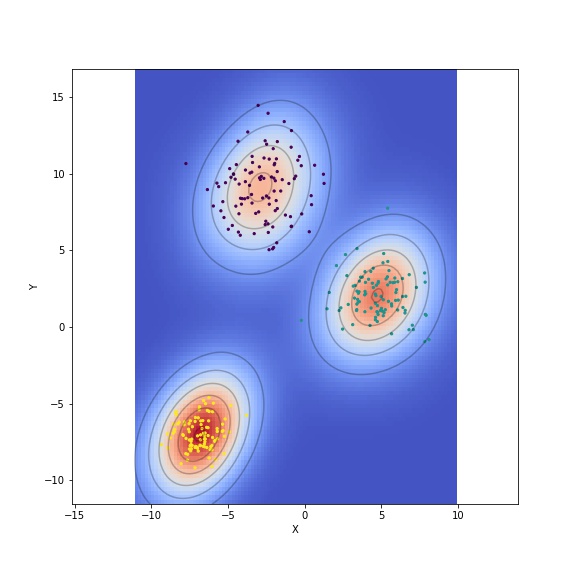} }}%
    \qquad
    \subfloat[Density model with level set]{{\includegraphics[width=0.25\textwidth]{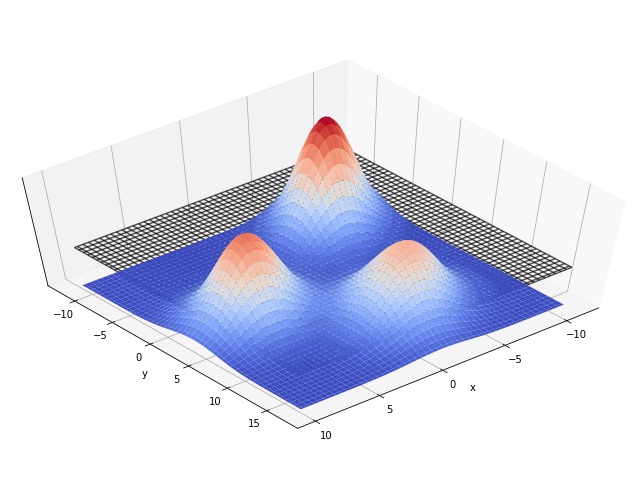} }}%
    \caption{Dataset with its corresponding contour lines and density model}%
    \label{fig:dbscanSurf}%
\end{figure}

One common assumption in Machine Learning is that the data points $x_i \in \mathbb{R}^d$ are  i.i.d. samples from a unknown \textit{density function} $\mathcal{D}$. We can think of this density function as a hypersurface in $\mathbb{R}^{d+1}$. The peaks of this hypersurface corresponds to the denser parts of the space $\mathbb{R}^d$, i.e., where data points are more concentrated. This is illustrated in figure \ref{fig:dbscanSurf}. In the example shown, data is drawn from a GMM with three components. Figure \ref{fig:dbscanSurf}(a) shows the sampled data points. Figure \ref{fig:dbscanSurf}(b) shows the density contour lines. Figure \ref{fig:dbscanSurf}(c) shows the density function, and a hyperplane representing a level of density, yielding three clusters. 

\begin{figure}[h]
    \centering
    \includegraphics[width=0.45\textwidth]{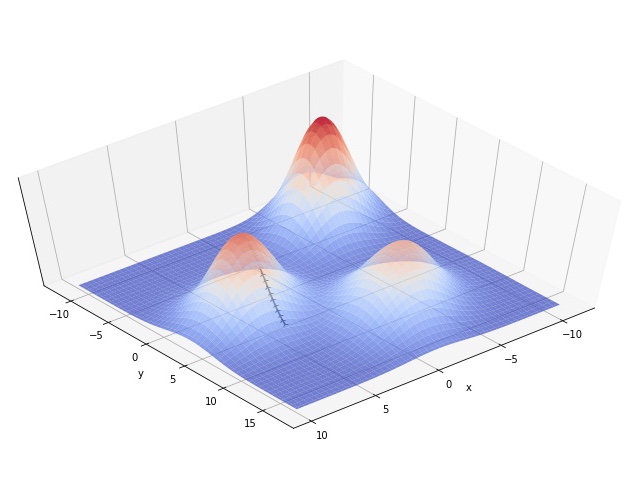}
    \caption{Example of a data point "climbing" the density function}
    \label{fig:dbscanClimb}
\end{figure}

We will use this idea to present a different way to find the core points in a dataset $X$ sampled from a density function $\mathcal{D}$. The basic idea is quite simple: every data point $x_i$ in $X$ will "climb the hill", trying to reach it closest peak (i.e., local maxima), as shown in image \ref{fig:dbscanClimb}, until the climbing point reaches a desired level of density (i.e., until it reaches a desired "altitude"). Because we don't know how the density function looks like, we will use the data itself to approximate the shape of the density function and the desired altitude.

In order to "climb the hill", we will follow a simple procedure proposed by Yizong in \cite{yizong}: given a dataset $X$, a data point $x_i$ and a radius $\epsilon$, they first compute the $\epsilon$-neighborhood $N_\epsilon(x_i) = \{x_j \in X | \norm{x_i - x_j} \leq \epsilon\}$, and then, using this local information, they compute the mean of the points in $N_\epsilon(x_i)$. This mean is the first stop in the path to the peak, and the algorithm iterates until the point reaches the desired altitude, for every data point. A detailed explanation on why this simple procedure is guaranteed to climb the hill is given in \cite{comaniciu} and \cite{yizong}. The main idea is that this procedure estimates the gradient of the density function $\mathcal{D}$. To see this, consider the \textit{kernel density estimator} \cite{bishop, comaniciu} $f_{\kappa}(x)$:

\begin{equation}
f_{\kappa}(x) = \frac{1}{nh^d}\sum_{i=1}^n\kappa\Bigg(\norm{\frac{x-x_i}{h}}^2\Bigg)
\end{equation}

where $h^D$ is the volume of a hypercube of side $h$ in a $d$ dimensional space. It can be seen that, as its name suggest, $f_{\kappa}(x)$ is an estimator of the value of the density $\mathcal{D}$ at the point $x$. This estimation is done using the sampled data points $x_i$. Given the estimator $f_{\kappa}(x)$, we can compute its gradient w.r.t. $x$

\begin{align*}
\nabla f_{\kappa}(x) &= \frac{2}{nh^{d+2}}\sum_{i=1}^n(x - x_i)\kappa'\Bigg(\norm{\frac{x-x_i}{h}}^2\Bigg) \\
&= \frac{2}{nh^{d+2}} \Bigg[\sum_{i=1}^n\kappa'\Bigg(\norm{\frac{x-x_i}{h}}^2\Bigg) \Bigg]\Bigg[x - \frac{\sum_{i=1}^nx_i\kappa'\big(\norm{\frac{x-x_i}{h}}^2\big)}{\sum_{i=1}^n\kappa'\big(\norm{\frac{x-x_i}{h}}^2\big)}\Bigg] \\
\end{align*}

where $\kappa'()$ is the derivative of $\kappa()$. We can observe that the first term of $\nabla f_{\kappa}(x)$ is proportional to $f_{\kappa'}(x)$ (where $f_{\kappa'}(x)$ is a kernel density estimator using $\kappa'()$ as a kernel instead of $\kappa()$). The second term is the (negative) \textit{mean shift} \cite{comaniciu}:

\begin{equation}
m(x) = x - \frac{\sum_{i=1}^nx_i\kappa'\big(\norm{\frac{x-x_i}{h}}^2\big)}{\sum_{i=1}^n\kappa'\big(\norm{\frac{x-x_i}{h}}^2\big)}
\end{equation}

which is the difference between $x$ and a weighted mean of the data points $x_i$. Using the last two equations

\begin{equation}
\nabla f_{\kappa}(x) = \frac{2}{h^2}f_{\kappa'}(x)m(x)
\end{equation}

yielding

\begin{equation}
m(x) = \frac{h^2}{2}\frac{\nabla f_{\kappa}(x)}{f_{\kappa'}(x)}
\end{equation}

This last expression shows that the mean shift vector $m(x)$ is proportional to the gradient of the kernel density estimator $f_{\kappa}(x)$ and therefore, it is an estimate of the gradient of $\mathcal{D}$ at the point $x$. In order to climb the hill, we want to move in a direction proportional to $m(x)$:

\begin{align*}
x_i^{new} &= x - m(x) \\
&= \frac{\sum_{i=1}^nx_i\kappa'\big(\norm{\frac{x-x_i}{h}}^2\big)}{\sum_{i=1}^n\kappa'\big(\norm{\frac{x-x_i}{h}}^2\big)} \\
&= \frac{\sum_{i=1}^nx_i\kappa_H(x, x_i)}{\sum_{i=1}^n\kappa_H(x, x_i)} \\
&= \frac{\sum_{x_j \in N_\epsilon(x_i)}x_j}{|N_\epsilon(x_i)|}
\end{align*}

where we have set $h = 1$ and used the Heaviside kernel $\kappa_H()$ as $\kappa'()$ in the third equality.

Algorithm \ref{algo:DBSCAN_MS} shows DBSCAN with this modification. Only the first step is different, the rest of the steps are the same. For each data point $x_i$, the climbing procedure iterates until $|N_\epsilon(x_i)| \geq minPts$, i.e., until it has reached the desired altitude, or until $x_i^{new} = x_i$. Observe that we need to create a copy of the dataset $X$, because we need to perform the climbing w.r.t. the original (and static) dataset $X$. Also, observe that the last step (\textbf{Outliers identification step}) is not needed anymore: after the climbing procedure, every data point in $X_{climb}$ is a core point because it has climbed until a desired level of density. If some $x_i$ in $X_{climb}$ couldn't climb to this altitude, it means it is an outlier, and the \textbf{Connected components step} will find it as a size-one cluster.

\begin{algorithm}
\DontPrintSemicolon 
\KwIn{Dataset $X$, radius $\epsilon$, and minimum number of points $minPts$}
\KwOut{Dense-connected clusters and outliers in $X$}

\textbf{Copy step}. Make a copy of the original dataset $X$: \\

\begin{equation*}
X_{climb} \gets X
\end{equation*}

\textbf{Core points identification step}. For every $x_i$ in $X_{climb}$: \\ 

\While{$|N_\epsilon(x_i)| < minPts$ or $x_i^{new} = x_i$}{

\begin{equation*}
x^{new}_i \gets \frac{\sum_{x_j \in N_\epsilon(x_i)}x_j}{|N_\epsilon(x_i)|}
\end{equation*}
}

\textbf{Graph step}. Create a graph $G = (\mathcal{V}, \mathcal{E})$ such that, for every core point in $X_{climb}$, there is a corresponding "core"node in $\mathcal{V}$, and add an edge to $\mathcal{E}$ between this node and all nodes in its $\epsilon$-neighborhood $N_\epsilon(x_i)$ \\

\textbf{Connected components step}. Find the connected components of graph $G$ \\

\Return{Connected components (and outliers) in $G$}\;
\caption{DBSCAN, Climbing hill version}
\label{algo:DBSCAN_MS}
\end{algorithm}

We will find how this version of DBSCAN is related to the Mean Shift algorithm in the next section.

\section{Mean Shift (MS)}

There are different versions of the Mean Shift algorithm. The common point between them is the climbing procedure, which is very similar to the climbing procedure mentioned in the previous section. The main difference between them is the clustering strategy: different versions of the algorithm use different criteria to create clusters. In this document, the clustering criteria will be the same criteria as the climbing hill version of DBSCAN, presented in the previous section.

The main and only difference between the climbing version of DBSCAN presented in the last section and the Mean Shift algorithm, is that, during the climbing procedure, points in the Mean Shift algorithm doesn't stop to climb once a desired level of density is reached: the climbing points keeps climbing until they reach their closest local maxima, as shown in algorithm \ref{algo:MS}. This simple difference produces (most of the times) very different results to the ones found by DBSCAN.

\begin{figure}[h]
    \centering
    \includegraphics[width=0.45\textwidth]{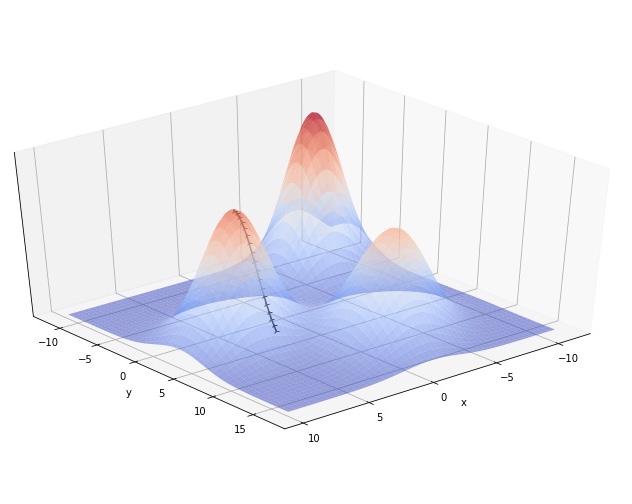}
    \caption{In MS, a data point keeps "climbing" until it reachs a "peak"}
    \label{fig:MSClimb}
\end{figure}

\subsection{Mean Shift complexity}

Copying the original dataset $X$ in the \textbf{Copy step} takes $O(nd)$ operations. Identifying the core points in $X$ using the climbing hill procedure can take up to $O(n^2dT)$ operations, because we need to repeatedly (at most $T$ times) compute the distance between $x_i$ in $X_{climb}$ and all points in $X$, for every point $x_i$ in $X_{climb}$. Constructing the graph in the \textbf{Graph step} takes $O(n_{core}^2)$ operations, because that is the maximum possible number of edges. Computing the connected components of $G$ in the \textbf{Connected components step} using DFS can take up to $O(n_{core}^2)$ operations. Therefore, the complexity of Mean Shift is $O(n^2dT)$. This complexity and other important properties of Mean Shift are summarized in table \ref{tab:MS}.


\begin{algorithm}
\DontPrintSemicolon 
\KwIn{Dataset $X$, radius $\epsilon$}
\KwOut{Dense-connected clusters and outliers in $X$}

\textbf{Copy step}. Make a copy of the original dataset $X$: \\

\begin{equation*}
X_{climb} \gets X
\end{equation*}

\textbf{Core points identification step}. For every $x_i$ in $X_{climb}$: \\ 

\While{$x_i$ hasn't reached a peak  or $x_i^{new} = x_i$}{

\begin{equation*}
x^{new}_i \gets \frac{\sum_{x_j \in N_\epsilon(x_i)}x_j}{|N_\epsilon(x_i)|}
\end{equation*}
}

\textbf{Graph step}. Create a graph $G = (\mathcal{V}, \mathcal{E})$ such that, for every core point in $X_{climb}$, there is a corresponding "core"node in $\mathcal{V}$, and add an edge to $\mathcal{E}$ between this node and all nodes in its $\epsilon$-neighborhood $N_\epsilon(x_i)$ \\

\textbf{Connected components step}. Find the connected components of graph $G$ \\

\Return{Connected components (and outliers) in $G$}\;
\caption{Mean Shift}
\label{algo:MS}
\end{algorithm}

\begin{table}[ht]
\centering
\caption{Mean Shift algorithm}
\begin{tabular}[t]{lc}
\hline
&MS\\
\hline
Number of hyperparameters & 1 \\
Hyperparameters & bandwith (radius) $\epsilon$  \\
Time complexity & $O(n^2dT)$ \\
Outlier detection? & Yes \\
Type of assignment & hard \\
distance/similarity measure & Heaviside kernel $\kappa_H()$ (See Mean Shift section for details) \\
\hline
\end{tabular}
\label{tab:MS}
\end{table}

\newpage

\section{Conclusions and future work}

This document presents the relationship between different clustering algorithms. A connection between DBSCAN and Mean Shift is stablished: DBSCAN can be viewed as a climbing procedure that stops once a data point has reached a given value of density. Mean shift follows a similar procedure, but its stopping criteria is different, it stops until a data point has reached its closest local maxima. Some of the key differences between the algorithms presented in this document are:

\begin{itemize}
    \item \textbf{The distance/similarity measure used}. This choice have a huge impact on the flexibility of the algorithm, i.e, what possible shapes can the algorithm find
    \item \textbf{The optimization method}. This choice determines the set of possible solutions (clusterings) that an algorithm can find. For instance, the solutions found by the EM algorithm and the eigendecomposition method are usually different, and each optimization methods has its advantages and disadvantages. EM is usually faster than the eigendecomposition method, but it usually returns a local optimum. On the other hand, the eigendecomposition method is guaranted to find the global optimum of the related matrix problem. For example, in section 4, the eigendecomposition method finds the global optimum of the \underline{relaxed} matrix version of the weighted kernel $k$-means problem
\end{itemize}

\begin{figure}[h]
    \centering
    \includegraphics[width=0.8\textwidth]{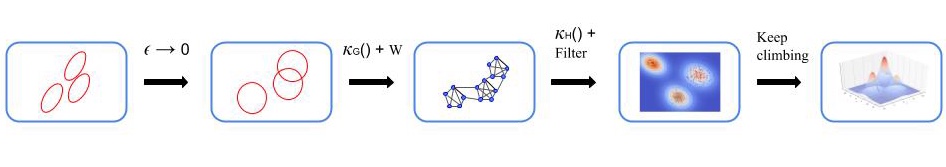}
    \caption{Relationship between GMM, $k$-means, SC, DBSCAN and MS}
    \label{fig:finalFig}
\end{figure}

These choices have a strong impact in the trade-off between the flexibility and the time complexity of each algorithm. Table \ref{tab:algComp} presents a summary of some of the main features of each algorithm. Figure \ref{fig:finalFig} answers the question in figure \ref{fig:algs}:

\begin{itemize}
    \item Setting all the covariance matrices equal to $\epsilon I$ and considering the limit $\epsilon \to 0$ take us from GMM to $k$-means
    \item Adding flexibility to $k$-means via a Gaussian kernel $\kappa_G()$ and introducing a weight $w_i$ for every data point yields a Spectral Clustering algorithm
    \item Using a different kernel $\kappa_H()$ and introducing a filtering step take us from Spectral Clustering to DBSCAN
    \item Climbing to the peak instead of stopping at a certain level take us from DBSCAN to Mean shift
\end{itemize}

\begin{table}[h]
\centering
\caption{Algorithms comparison}
\begin{tabular}[t]{lccccc}
\hline
&GMM &$k$-means &SC &DBSCAN &MS\\
\hline
Number of hyperparameters & 1 & 1 & 2 & 2 & 1 \\
Hyperparameters & $k$ & $k$ & $k,\sigma$ & $\epsilon,minPts$ & $\epsilon$ \\
Time complexity & $O(nd^2kT + d^{2.807}kT)$ & $O(ndkT)$ & $O(n^2d)$ & $O(n^2d)$ & $O(n^2dT)$ \\
Outlier detection & No & No & No & Yes & Yes \\
Type of assignment & soft & hard & hard & hard & hard \\
distance/similarity measure & Mahalanobis & Euclidean & Gaussian & Heaviside & Heaviside \\
\hline
\end{tabular}
\label{tab:algComp}
\end{table}

\subsection{Future work}

In future work, we would like to investigate what might be the relationship between some of the algorithms presented and some other clustering algorithms, like \textit{Hierarchical clustering} algorithms or some other non-parametric methods, like \textit{Latent Dirichlet Allocation} (LDA) \cite{blei} or the \textit{Chinese Restaurant Process} (CRP) \cite{pitman}. Similarly, it would be interesting to explore the possible relationship between some of the algorithms presented and some recently developed neural-based unsupervised learning algorithms, like the \textit{Stacked Capsule Autoencoders} \cite{hinton}, \textit{Deep InfoMax} (DIM) \cite{bengio}, \textit{Invariant Information Clustering} (IIC) \cite{ji}, or \textit{Deep Embedded Clustering} (DEC) \cite{xie}.

\end{document}